\documentclass{article}  

\usepackage{microtype}
\usepackage{graphicx}
\usepackage{subcaption}
\usepackage{lscape}
\usepackage{booktabs} 
\usepackage[table]{xcolor}

\usepackage{hyperref}

\usepackage[accepted]{icml2024}

\usepackage{amsmath}
\usepackage{amssymb}
\usepackage{mathtools}
\usepackage{amsthm}
\usepackage{adjustbox}
\usepackage{changepage} 

\usepackage[capitalize,noabbrev]{cleveref}

\usepackage[textsize=tiny]{todonotes}

\usepackage{enumitem} 
\definecolor{bluecite}{HTML}{0875b7}
\definecolor{Gray}{gray}{0.96}
\usepackage{longtable}
\usepackage{tabularx}
\usepackage{todonotes}
\usepackage[normalem]{ulem}

\icmltitlerunning{Should we be going MAD? A Look at Multi-Agent Debate Strategies for LLMs}

\begin{document}

\twocolumn[
\icmltitle{Should we be going MAD? \\ A Look at Multi-Agent Debate Strategies for LLMs}



\icmlsetsymbol{equal}{*}

\begin{icmlauthorlist}
\icmlauthor{Andries Smit}{equal,yyy}
\icmlauthor{Nathan Grinsztajn}{equal,yyy} 
\icmlauthor{Paul Duckworth}{equal,yyy}
\icmlauthor{Thomas D.\ Barrett}{yyy}
\icmlauthor{Arnu Pretorius}{yyy}
\end{icmlauthorlist}

\icmlaffiliation{yyy}{InstaDeep, London UK}
\icmlcorrespondingauthor{Paul Duckworth}{p.duckworth@instadeep.com}

\icmlkeywords{Machine Learning, ICML}

\vskip 0.3in
]



\printAffiliationsAndNotice{\icmlEqualContribution} 

\begin{abstract}

Recent advancements in large language models (LLMs) underscore their potential for responding to inquiries in various domains. However, ensuring that generative agents provide accurate and reliable answers remains an ongoing challenge. In this context, multi-agent debate (MAD) has emerged as a promising strategy for enhancing the truthfulness of LLMs.
%
We benchmark a range of debating and prompting strategies to explore the trade-offs between cost, time, and accuracy. Importantly, we find that multi-agent debating systems, in their current form, do not reliably outperform other proposed prompting strategies, such as self-consistency and ensembling using multiple reasoning paths. However, when performing hyperparameter tuning, several MAD systems, such as Multi-Persona, perform better. This suggests that MAD protocols might not be inherently worse than other approaches, but that they are more sensitive to different hyperparameter settings and difficult to optimize.
%
We build on these results to offer insights into improving debating strategies, such as adjusting agent agreement levels, which can significantly enhance performance and even surpass all other non-debate protocols we evaluated.  We provide an open-source repository to the community with several state-of-the-art protocols together with evaluation scripts to benchmark across popular research datasets.
\end{abstract}

\section{Introduction} \label{sec:introduction}

Truthful question-answering assistants require a plethora of skills that until recently were considered out-of-reach of generative language models. Such agents require advanced natural language reading comprehension, along with accurate recall and manipulation of expert or technical knowledge. Following the increase in performance and popularity of large language models (LLMs) \citep{brown2020language, openAI2023gpt4}, there is a growing hope that these models could assist humans in various domains like medicine \citep{lievin2022can, han2023medalpaca, nori2023medprompt, wang2023clinicalgpt, singhal2023medpalm}, education \citep{pardos2023learning, kumar2024learning}, and law \citep{lai2023law}.
As a way to improve the reasoning abilities of these systems, prompting strategies have received significant attention, starting with few-shot~\citep{brown2020language}, and chain-of-thought (CoT)~\citep{wei2022chain, kojima2022large}. 
To further improve performance, a wide variety of strategies have been proposed to use interactive reasoning between multiple LLMs, by either generating answers in parallel to maintain a form of self-consistency~\citep{wang2022self}, or promoting models to simulate debate. 
These multi-agent approaches have recently seen an uptake in applications, e.g. language generation~\citep{chan2023chateval}, 
machine translation and arithmetic reasoning~\citep{liang2023encouraging}, mathematical and strategic reasoning~\citep{du2023improving},
negotiation and bargaining~\citep{fu2023improving},
and notably, medical Q\&A~\citep{anil2023palm, singhal2023medpalm, nori2023medprompt}.
%
How to best utilize multiple agents for effective interactive reasoning is a prescient research question. However, to the best of our knowledge, there is no work comparing strategies, and there is no consensus for selecting one strategy over another. 

In this paper, we benchmark various prompting strategies, which include multi-agent debate (MAD), for answering multiple-choice questions across a wide range of domains. We explore the impact on, and trade-offs between, critical factors such as factual accuracy, time and cost.  
We provide an open-source suite of single-agent and MAD implementations for the research community to build upon, with a unified API to easily build and evaluate MAD systems. Finally, we demonstrate that by utilizing specific prompting strategies, LLMs exhibit improved reasoning abilities. Concretely, we provide a novel debate prompting strategy able to modulate the level of agreement between agents during a debate and improve upon the state-of-the-art for Q\&A for a given model class. 


\section{Multi-Agent Debate} \label{sec:mad-strategies}

Current state-of-the-art models for Q\&A are dominated by generative LLMs. To make them more truthful and reliable, they are often fine-tuned for specific use cases. In medicine, for example, such models include 
Med-PaLM~\citep{singhal2022large}, 
Med-PaLM2~\citep{singhal2023medpalm}, 
MedAlpaca~\citep{han2023medalpaca}, 
Galactica~\citep{taylor2022galactica},
ClinicalGPT~\citep{wang2023clinicalgpt} and Medprompt ~\citep{nori2023medprompt}. 
Furthermore, many single-agent prompting strategies have been investigated in the context of medical Q\&A. 
For example, \citet{lievin2022can} 
applied CoT reasoning on top of Instruct GPT-3~\citep{ouyang2022training}, and achieves noticeable performance improvements.

Recently, several MAD strategies have been proposed to improve upon the enhanced reasoning capabilities of single-agent prompting methods 
leading to improved performance on challenging natural language tasks~\citep{du2023improving, liang2023encouraging, chan2023chateval}. 
Likewise, Generative agents~\citep{park2023generative}, multi-persona~\citep{wang2023unleashing}, and CAMEL~\citep{li2023camel} study the behaviour of agents taking on different roles or personas within multi-agent interactions. 
One major reason why debate strategies can be an effective tool is the ability of LLMs to adapt to additional information given in-context~\citep{zhang2023and}. 
This facilitates multiple LLMs to participate in multi-agent and/or multi-round debates entirely using in-context prompting. That is, the agents adapt their behaviour based on information provided by other agents at inference time, with no gradient-based parameter updates being required. 

In our study, we investigate several prompting strategies for Q\&A. Whilst we note that some of these strategies were introduced specifically for the medical Q\&A domain, each provides novel perspectives on how to utilize multiple collaborative agents. 
We briefly introduce each strategy here. 

\textbf{Society of Minds (SoM)} ~\citet{du2023improving} propose a MAD approach where multiple agents each provide their answers to each other in order to effectively collaborate. Optionally, answers are summarized before being added to the history that is available to the agents in future rounds.

\textbf{Multi-Persona} ~\cite{liang2023encouraging} propose a MAD strategy to encourage divergent agent outcomes via prompting different personalities, i.e.\ an affirmative agent (angel) and a negative agent (devil) each provide an answer to a judge agent who manages the process and obtains a final solution. The judge has additional agency to end the debate early if it is satisfied with the answers provided. 

\textbf{ChatEval} ~\cite{chan2023chateval} propose three MAD modes: 
(1) \emph{one-on-one}, where each agent answers the provided question in turn, and each agent is provided with the history of all previous agents' answers; 
(2) \emph{simultaneous-talk}, where agents asynchronously generate responses in each round to nullify the effects of agent order; and
(3) \emph{simultaneous-talk-with-summarizer}, which additionally summarizes answers provided in each round and overwrites the history available to all agents in future rounds. 

\textbf{Self-consistency}~\cite{wang2022self} samples multiple reasoning paths given a fixed prompt and selects the most frequent answer. 
Whilst this is not a debate per se, as samples are rolled out independently, it relies on multiple API calls so we distinguish it from the single agent case that uses a single API call.

\textbf{Ensemble Refinement (ER)} ~\cite{singhal2023medpalm} extends self-consistency. After multiple reasoning paths are sampled, a second stage concatenates them into a list of \textit{student reasonings}, after which multiple rounds of aggregation are performed conditioned on this list. 

\textbf{Medprompt} ~\citep{nori2023medprompt} enhances AI-driven medical question answering by combining few-shot exemplar selection with a two-stage process: preprocessing correct GPT-4 training set responses as few-shot examples, and an inference step employing a k-Nearest Neighbours (kNN) lookup and ensemble refinement for answer generation. This approach ensures high accuracy and reliability by using only the most relevant and validated examples. Medprompt's effectiveness is demonstrated in its superior performance on the MedQA benchmark, showcasing its potential for medical diagnostics. In this work, we do not employ the kNN approach as it requires a training and test set split. We believe this might provide Medprompt an unfair advantage in this evaluation as other protocols do not use a training set. Therefore, we only implement subcomponents of the full Medprompt, specifically, question randomization with few-shot chain-of-thought ensembling~\cite{nori2023medprompt}.

\begin{table*}[t!]
 \centering
 \rowcolors{2}{Gray}{white}
 \begin{adjustbox}{width=\linewidth,center}
 \begin{tabular}{lcccccc}
    \textbf{System} & \textbf{\begin{tabular}[c]{@{}c@{}}Flexible \\ Round Numbers\end{tabular}} & \textbf{Judge} & \textbf{Summarizer} & \textbf{\begin{tabular}[c]{@{}c@{}}Sequential \\ Interactions\end{tabular}} & \textbf{\begin{tabular}[c]{@{}c@{}}Multi API \\ Calls\end{tabular}} & \textbf{\begin{tabular}[c]{@{}c@{}}Asymmetric \\ Agents\end{tabular}} \\
    \midrule 
     Medprompt                &              &            &            & \checkmark & \checkmark &            \\
    Society of Minds         & \checkmark   &            &            & \checkmark & \checkmark &            \\
    Ensemble Refinement      &              & \checkmark & \checkmark & \checkmark & \checkmark &            \\
    ChatEval                 &  \checkmark  & \checkmark & \checkmark & \checkmark & \checkmark &            \\
    Self-consistency         &              &            &            &            & \checkmark &            \\
    Single-agent             &              &            &            &            &            &            \\
    Solo Performance Prompting                      & \checkmark   &            &            & \checkmark &            & \checkmark \\
    Multi-Persona            & \checkmark   & \checkmark &            & \checkmark & \checkmark & \checkmark \\
   
    \bottomrule 
 \end{tabular}
\end{adjustbox}
 \caption{Feature comparison of various debating systems. The presence of a feature is indicated by a checkmark.} \label{tab:feature_comparison}
\end{table*}

We summarize the features of these various systems in Table~\ref{tab:feature_comparison}.
Each strategy determines the high-level debate prompts and how the agents share answers and histories to collaborate. However, in each case, there are multiple possible agent-level prompts available, including: 
(1)~zero-shot Q\&A prompt,
(2) zero-shot chain-of-thought (CoT)~\citep{kojima2022large}, 
(3) few-shot examples~\citep{brown2020language} which provides five Q\&A examples but no reasoning,
(4) Solo Performance Prompting (SPP) (or Multi-persona self-collaboration)~\citep{wang2023unleashing} which utilizes a single agent that mimics an internal debate, and
(5) few-shot chain-of-thought (FS-CoT)~\citep{wei2022chain} 
 which combines step-by-step reasoning steps, along with five Q\&A examples and explanations\footnote{The few-shot Q\&A examples with explanations are provided for each medical dataset in~\citet{singhal2023medpalm}. The step-by-step explanations were generated by a panel of qualified clinicians who identified the best examples and crafted the few-shot prompts as part of the Med-PaLM project.}.
We provide the complete lists of all debate and agent-level prompts in Appendices \ref{app:debate_prompts} and~\ref{app:agent_prompts}.

\begin{figure*}[t!]
  \centering
    {\includegraphics[width=0.54\textwidth]{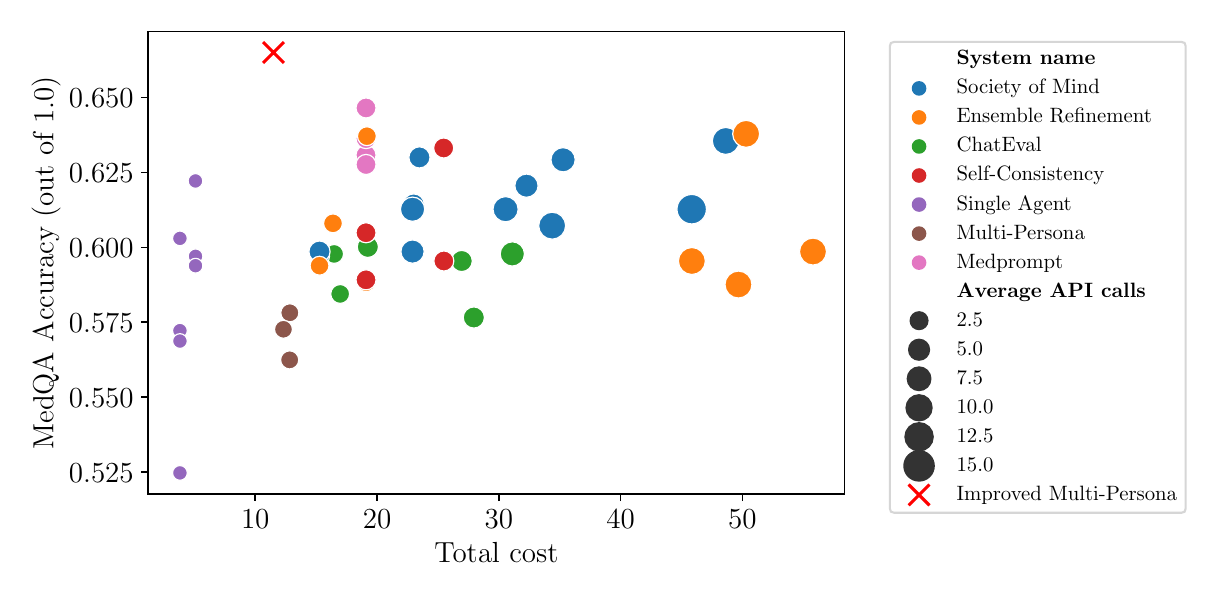}}
    {\includegraphics[width=0.41\textwidth]{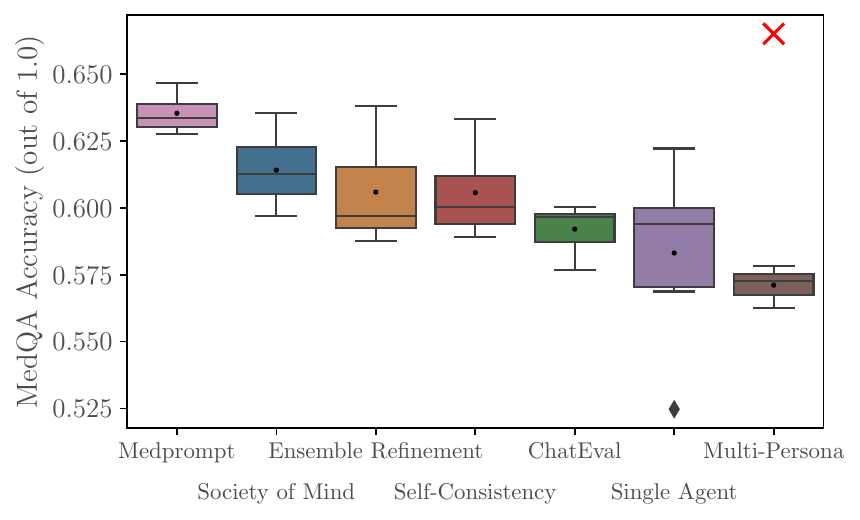}}
    \caption{\textit{Benchmark of experiment configurations on MedQA dataset.} \textbf{Left}: Accuracy vs average cost (\$) per question. The size of the dots reflects the average number of API calls required per question. 
    \textbf{Right}: Summarizes accuracy grouped by strategy, sorted by average performance (black dot).
    The {\color{red}X} represents improved performance using our proposed \emph{agreement modulation}, described in Section \ref{sec:agreement_modulation}.} 
    \label{fig:medqa_scatter}
\end{figure*}

\section{Experiments} \label{experiments}

As base agents for the MAD implementations, we utilize GPT-3~\citep{brown2020language} with the 3.5-turbo engine. GPT-3 is a large-scale transformer-based generative LLM~\citep{vaswani2017attention, kaplan2020scaling} available via API calls. Although not the current most powerful model available via API, we selected GPT-3.5 for its optimal balance between performance and cost efficiency, crucial for conducting a broad range of experiments, whereas more powerful models would've been prohibitively expensive. However, we informally note that some of our preliminary experiments with more advanced models like GPT-4 suggested similar findings, although these tests were limited to a smaller subset of data. 

We evaluate each system using seven datasets: three medical datasets, and three more general datasets requiring reasoning. 
For the medical tasks, we follow the evaluation protocol in Med-PaLM2~\citep{singhal2023medpalm} and evaluate the above strategies on the following multiple-choice question-answering datasets: 
\begin{itemize}[leftmargin=18pt]
    \item \textbf{MedQA}~\citep{jin2021disease} comprising of 1273 general medical knowledge questions from the US medical licensing exam (USMLE). Each question has 4-5 answer choices. In this work, we focus on the 4-answer version of MedQA as used to evaluate Med-PaLM2~\cite{liang2023encouraging} and Medprompt~\cite{nori2023medprompt}.
 \item \textbf{PubMedQA}~\citep{jin2019pubmedqa} containing 500 open domain questions, context and answers. 
 \item \textbf{MMLU} (clinical topics only)~\citep{hendrycks2020measuring} consisting of 123 medical questions covering anatomy, clinical knowledge, college medicine, medical genetics, professional medicine, and college biology.
\end{itemize}

The other datasets are difficult reasoning tasks, based on the assumptions that MAD strategies might particularly shine on datasets requiring complex and commonsense reasoning and be better able to escape reasoning traps.

\begin{itemize}
\item \textbf{CosmosQA}~\citep{2019CosmosQA}: A dataset of 7,000 general knowledge questions focusing on commonsense-based reading comprehension. It requires understanding narratives and interpreting causes and effects not explicitly mentioned. We subsampled 500 questions for economic and time feasibility.
\item \textbf{CIAR} ~\citep{liang2023encouraging}: The Counter-Intuitive Arithmetic Reasoning (CIAR) dataset consists of 50 multiple-choice questions. It is designed to evaluate reasoning abilities of LLMs, challenging them to use slow, logical thinking over intuitive responses. We added at the end of the question the two suggested answers to turn it into a multiple-choice question-answering task. 
\item \textbf{GPQA}~\citep{rein2023gpqa}: This dataset contains 448 multiple-choice questions written by domain experts in biology, physics, and chemistry. It tests the limits of both human experts and AI systems, with questions that are `Google-proof' and demand high-level understanding and reasoning.
\item \textbf{Chess}~\citep{chessstatetracking}: This benchmark task focuses on the ability of AI systems to understand and track the state of a chess game based on natural language descriptions. The task is divided into subtasks created from real games played on Lichess and synthetic games generated by random search over the chess game tree. To account for the effect of context length, the subtasks are further divided into Short, Medium, and Long. Each subtask consists of 1,000 games. We use the Short setting in our study.
\end{itemize}

We measure additional agent-level and debate-level metrics (comprehensive lists of all additional metrics are provided in Appendices~\ref{app:all_debate_metrics} and \ref{app:all_agent_metrics}). 
Examples of agent-level metrics include whether an individual agent answered a given question correctly or not and the debate round in which it first provided the correct answer.  
Examples of debate-level metrics include whether \textit{any} agent involved in the debate provided a correct answer and 
whether the agents came to a consensus by the final round.

\paragraph{Results}
First, we attempt to see whether various prompting protocols affect performance significantly enough to justify their API costs. In Figure~\ref{fig:medqa_scatter}, we present a scatter plot of the results of each experiment on the MedQA dataset, while equivalent analyses for the other datasets are given in Appendix~\ref{app:results_overview}.
In the left panel, we show the accuracy vs cost (measured in USD), where the size of each point reflects the average number of API calls required per question (we also plot accuracy vs time, and accuracy vs average prompt length in Appendix~\ref{app:results_overview}). These variations, where applicable, include changing the number of agents, rounds, reasoning and aggregation steps, altering agent prompts, implementing round summarization, adjusting sampling parameters, and modifying other system-specific hyperparameters. For a full breakdown of the experimental configurations used for each system, see Appendix~\ref{app:all_exps}. 
In the right panel, we summarize accuracy over all configurations per strategy. 

We see the highest performing debating strategy from those introduced in Section~\ref{sec:mad-strategies} is SoM with multiple different configurations achieving 61\% on MedQA. However, the recently introduced Medprompt strategy~\cite{nori2023medprompt}, which does not use debate, performs the best overall. It also has lower costs associated with it.
Somewhat concerning for the domain of medical Q\&A, is that the single agent's performance, along with self-consistency, can be manipulated via prompts to achieve a high variance. 
We also note that Multi-Persona performs consistently about 7\% worse than the Medprompt strategy. We revisit this in Section \ref{sec:agreement_modulation}.

\begin{table*}[t!]
 \centering
  \begin{adjustbox}{width=\linewidth,center}
 \begin{tabular}{lccccccc}
    \textbf{System} & \textbf{MedQA} & \textbf{PubMedQA} & \textbf{MMLU} & \textbf{CosmosQA} & \textbf{CIAR} & \textbf{GPQA} & \textbf{Chess}\\
    \hline
    Medprompt            & \textbf{0.65} (0.63) & \textbf{0.77} (0.77) & 0.74 (0.73) & \textbf{0.48} (0.47) &0.54 (0.50) & 0.27 (0.25) & 0.32 (0.30)\\
    Society of Mind      & 0.64 (0.61) & 0.74 (0.71) & 0.73 (0.70) & 0.44 (0.39) &0.\textbf{56} (0.46) & 0.27 (0.25)& 0.26 (0.25) \\
    Ensemble Refinement  & 0.64 (0.60) & 0.74 (0.72) & 0.76 (0.74) & 0.45 (0.40) &0.48 (0.46) & 0.32 (0.26) & 0.32 (0.25)\\
    ChatEval             & 0.60 (0.60) & 0.75 (0.73) & 0.71 (0.69) & 0.45 (0.43) &0.48 (0.43) & 0.26 (0.25) & 0.32 (0.23)\\
    Self-Consistency     & 0.60 (0.60) & 0.74 (0.72) & \textbf{0.78} (0.75) & 0.46 (0.46) &\textbf{0.56} (0.52) & 0.24 (0.29) & 0.27 (0.21)\\
    Single Agent         & 0.60 (0.59) & 0.75 (0.70) & 0.76 (0.72) & 0.45 (0.43) &0.50 (0.50) & \textbf{0.33} (0.28) & 0.27 (0.18)\\
    Multi-Persona        & 0.58 (0.57) & 0.70 (0.69) & 0.72 (0.69) & 0.46 (0.42) & 0.52 (0.50) & 0.29 (0.29) & \textbf{0.33} (0.29)\\
    \bottomrule 
 \end{tabular}
\end{adjustbox}

     \caption{Best performance achieved by each system on various datasets. The highest score for each system on each dataset is listed, with the median performance over the datasets shown in parentheses.} \label{tab:best_config}
\end{table*}
\begin{figure*}[ht!]
\centering
\begin{subfigure}[b]{0.3\linewidth}
\includegraphics[width=\linewidth]{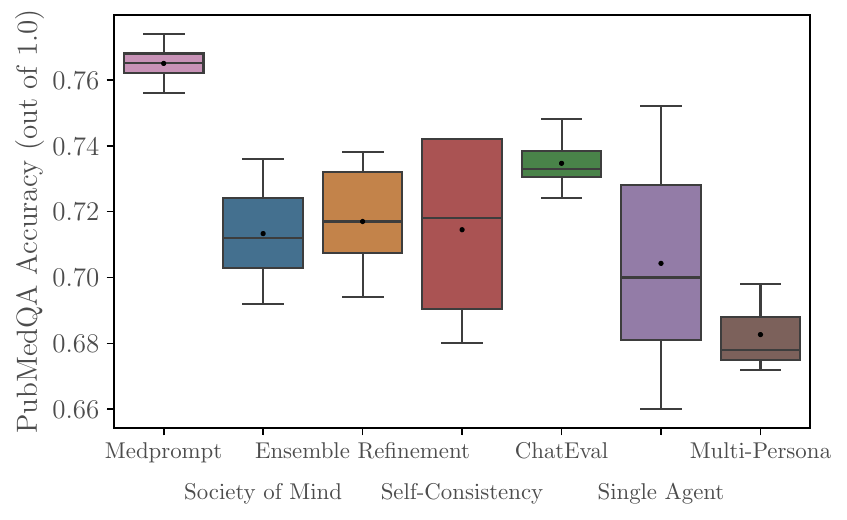}
\caption{PubMedQA Total Accuracy}
\end{subfigure}
\begin{subfigure}[b]{0.3\linewidth}
\includegraphics[width=\linewidth]{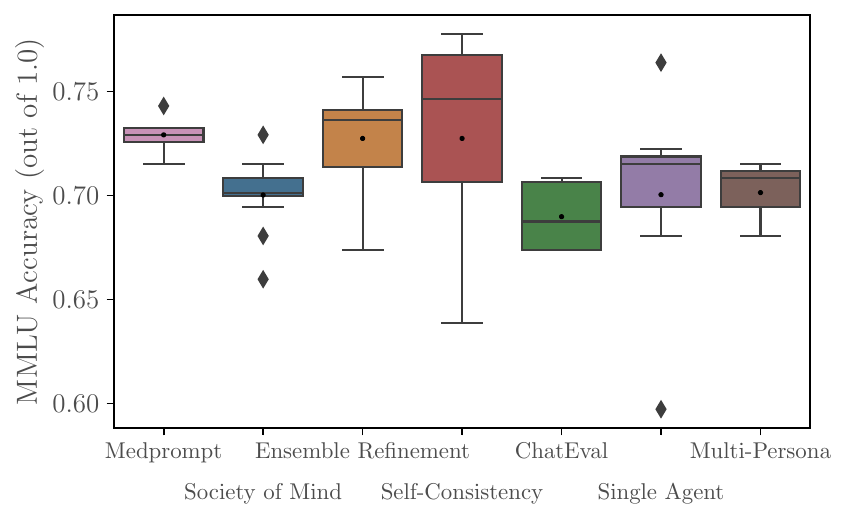}
\caption{MMLU Total Accuracy}
\end{subfigure}
\begin{subfigure}[b]{0.3\linewidth}
\includegraphics[width=\linewidth]{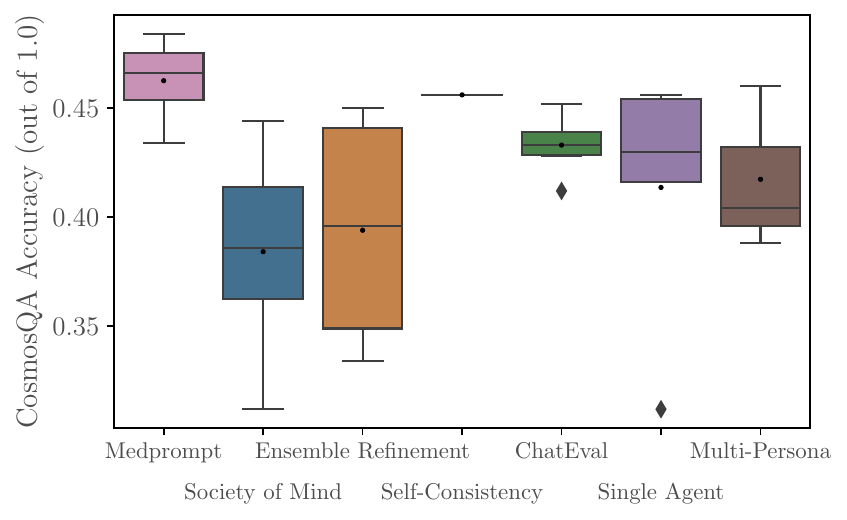}
\caption{CosmosQA Total Accuracy}
\end{subfigure}
\begin{subfigure}[b]{0.3\linewidth}
\includegraphics[width=\linewidth]{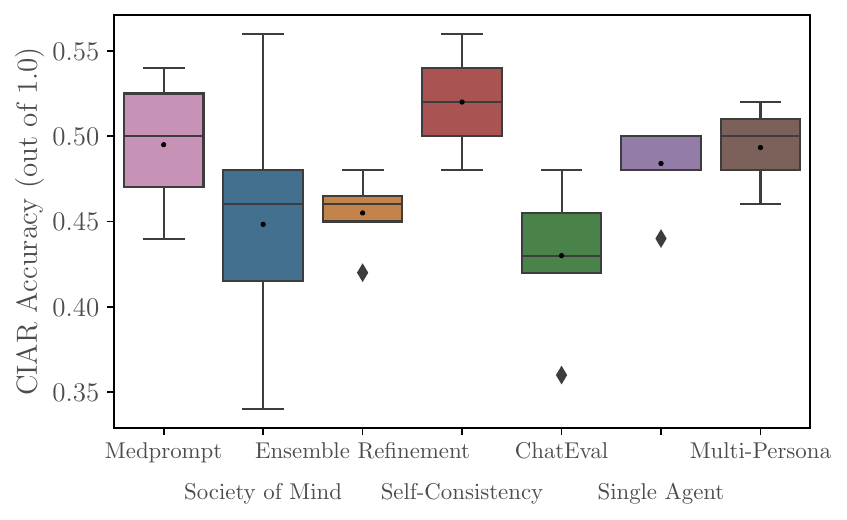}
\caption{CIAR Total Accuracy}
\end{subfigure}
\begin{subfigure}[b]{0.3\linewidth}
\includegraphics[width=\linewidth]{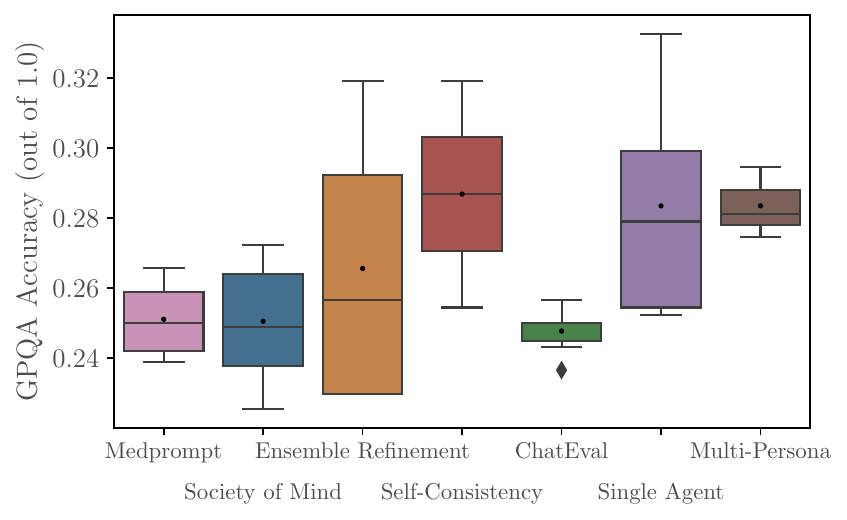}
\caption{GPQA Total Accuracy}
\end{subfigure}
\begin{subfigure}[b]{0.3\linewidth}
\includegraphics[width=\linewidth]{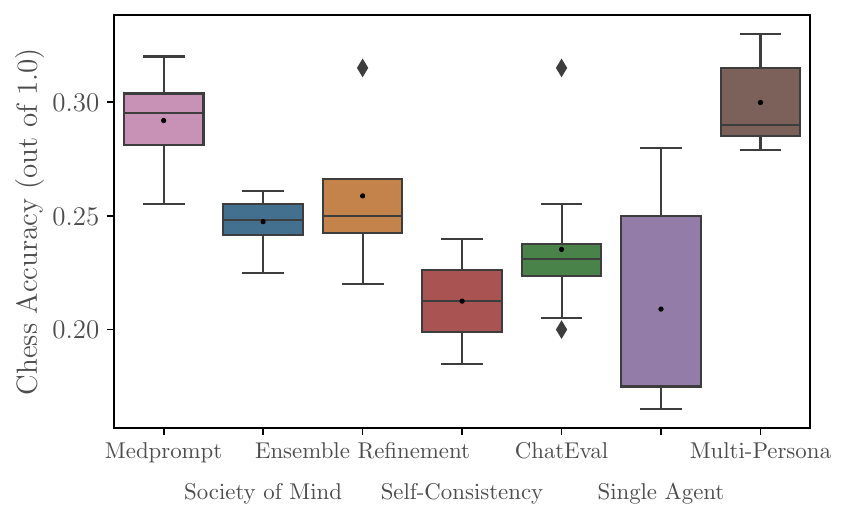}
\caption{Chess Total Accuracy}
\end{subfigure}
\caption{Boxplots illustrating the distribution of total accuracy across different datasets for each QA system. These plots demonstrate the variability and robustness of each system's performance.}
\label{fig:boxplots}
\end{figure*}

%


While there is a noticeable trend indicating improved performance with increased cost and API usage, our findings reveal that the relationship between performance and resource investment is complex and highly dependent on specific hyperparameters and dataset/system configurations. Notably, some MAD strategies do not consistently outperform more cost-effective non-debating approaches like Medprompt, even with optimized prompting. This suggests that, beyond a certain threshold, additional computing does not guarantee better results, and performance is both nuanced and contingent on the right combination of hyperparameters and system design. This variability underscores the need for further evaluations, as undertaken in the following section, to develop a clearer picture of these trends.



\paragraph{The utility of debate}
Next, we consider the utility of debate in prompting strategies. Recent protocols like Ensemble Refinement~\cite{singhal2023medpalm} and Medprompt~\cite{nori2023medprompt} have deviated away from debating altogether, raising questions about its efficacy. Recall that we include three standard medical datasets (MedQA, PubMedQA and MMLU), while also including three other reasoning datasets (CosmosQA, CIAR and GPQA). These datasets are constructed in such a way as to require small amounts of logical deduction to answer successfully. This might give debating protocols an advantage as there are multiple rounds of outputs between agents, which could be used to build on logical arguments before concluding.  We now evaluate all our system setups across these seven datasets. The scores (out of 1.0) for the best-performing configurations for each QA system can be found in Table \ref{tab:best_config}.

Table \ref{tab:best_config} indicates that no protocol dominates on all datasets. Medprompt seems to perform the best overall. Interestingly enough the debating protocols do not outperform the other protocols on the reasoning datasets. To further investigate this we provide box plots, in Figure \ref{fig:boxplots}, which show the distribution of performances for each QA system across the other datasets (MedQA's results can be found in Figure \ref{fig:medqa_scatter}). These boxplots help to visualize the variability and robustness of each system under different conditions.

The protocols with the most variance in performance seem to be Ensemble Refinement, Self-Consistency and single-agent implementations. Medprompt seems to perform consistently well, except on the GPQA dataset, with low variance. It seems that MAD protocols are more sensitive to hyperparameters. We investigate this further in the next section.

\paragraph{Is MAD simply sensitive to hyperparameters?}

To benchmark MAD's sensitivity to hyperparameter choices, we adopt a K-fold approach. We divide the medical and non-medical datasets into two groups. For each method and each dataset, we gather the hyperparameter combination which has the highest average accuracy on the two held-out datasets of the same category, and use this combination to compute the final accuracy.

\begin{figure*}[t!]
  \centering
    \includegraphics[width=0.45\textwidth]{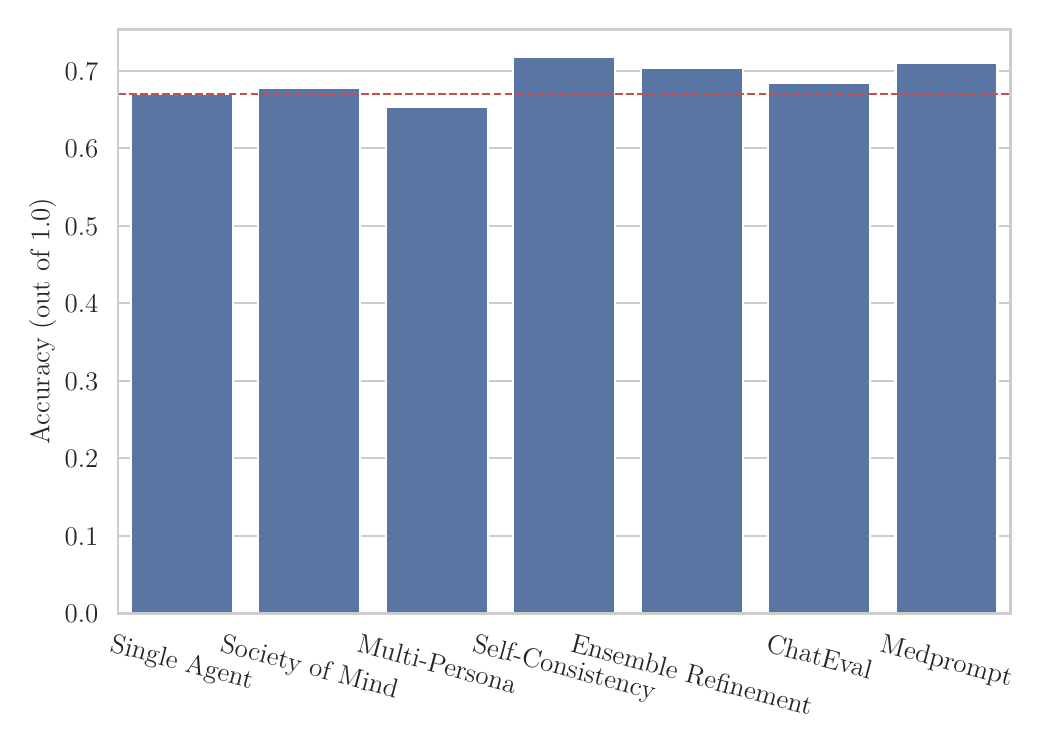}
    \includegraphics[width=0.45\textwidth]{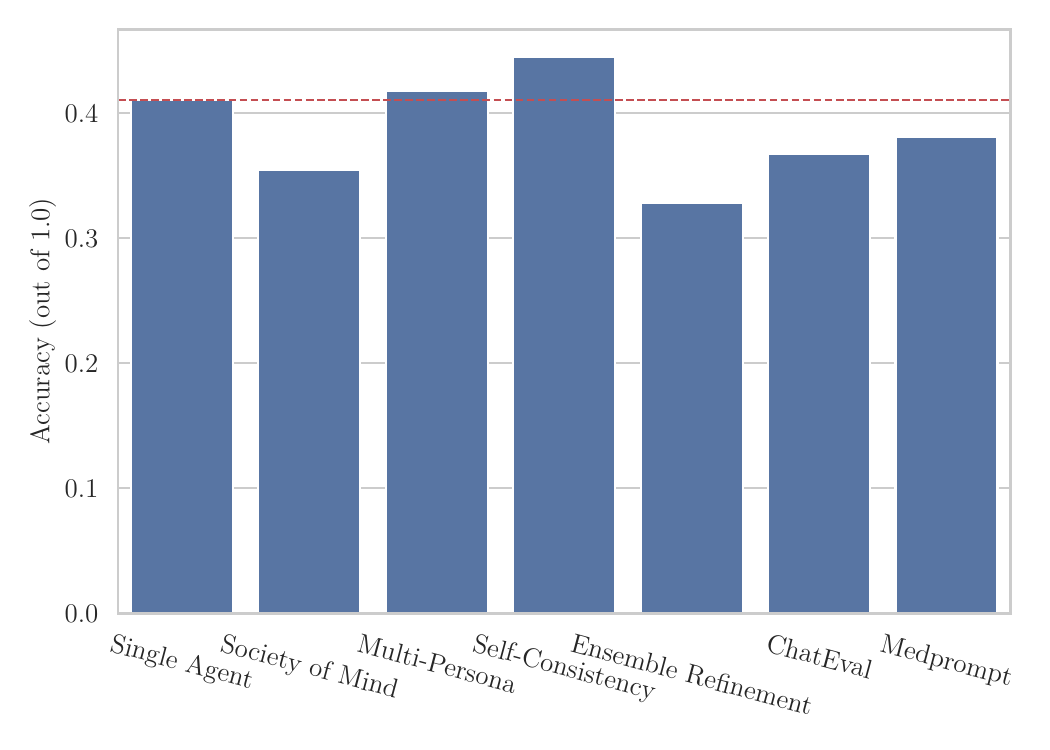}
    \caption{\emph{Comparing medical vs non-medical datasets}. The red line indicates the average performance of single-agent systems. \textbf{Left}: Accuracy averaged over the three medical datasets.
    \textbf{Right}: Accuracy averaged over the three non-medical datasets.}
    \label{fig:k_fold}
\end{figure*}

The results are presented in Figure~\ref{fig:k_fold}. Notably, within medical datasets, every evaluated system outperforms single-agent methods, with the exception of Multi-Persona. Self-Consistency and Medprompt emerge as the top performers. Conversely, in the context of non-medical datasets, single-agent methods surpass all evaluated systems except for Multi-Persona and Self-Consistency.

These findings indicate that hyperparameter choices significantly influence system performance, with optimal settings tending to be dataset-specific. Moreover, they underscore that superior performance of various protocols over a standard single-agent approach is not assured for new datasets. It appears that dataset-specific fine-tuning is crucial for maximizing performance.

\paragraph{Measuring the performance boost from debating}\label{sec:boost_debating}

\begin{figure}[htb!]
  \centering
    \includegraphics[width=0.45\textwidth]{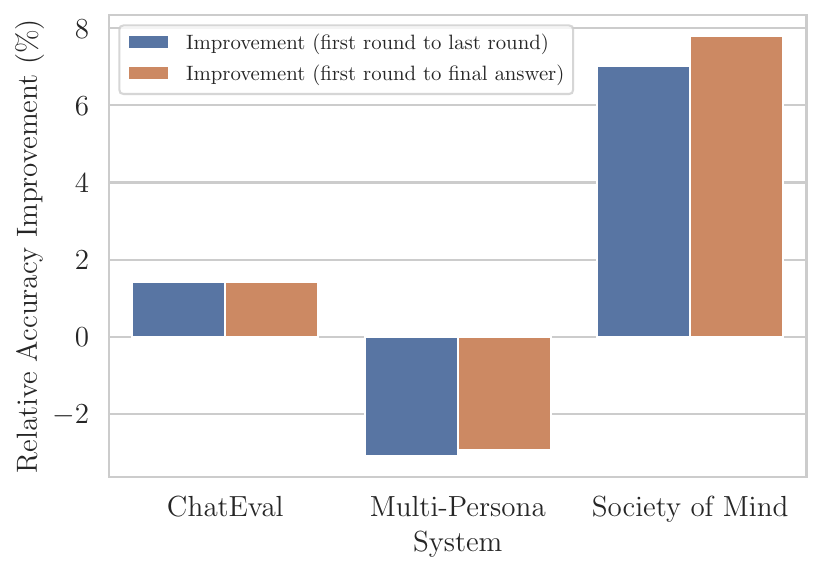}
    \caption{Relative accuracy improvements, averaged on all datasets, between the answer of the first agent given during the first and last round, as well as with the final answer.}
    \label{fig:relative_improvements}
\end{figure}

In Figure \ref{fig:relative_improvements}, we observe the relative improvement in accuracy achieved by the debate process compared to the initial answer provided by the first agent. Notably, the performance varies across the three systems: ChatEval exhibits a modest enhancement, Society of Mind shows a substantial increase, and Multi-Persona, intriguingly, leads to a decrease in performance. This divergence in outcomes, coupled with the findings in Table \ref{tab:best_config} which reveal only minor differences in final accuracy among these systems, suggests that the initial quality of responses in the first round may vary significantly.

\begin{figure*}[t!]
  \centering
    \includegraphics[width=0.45\textwidth]{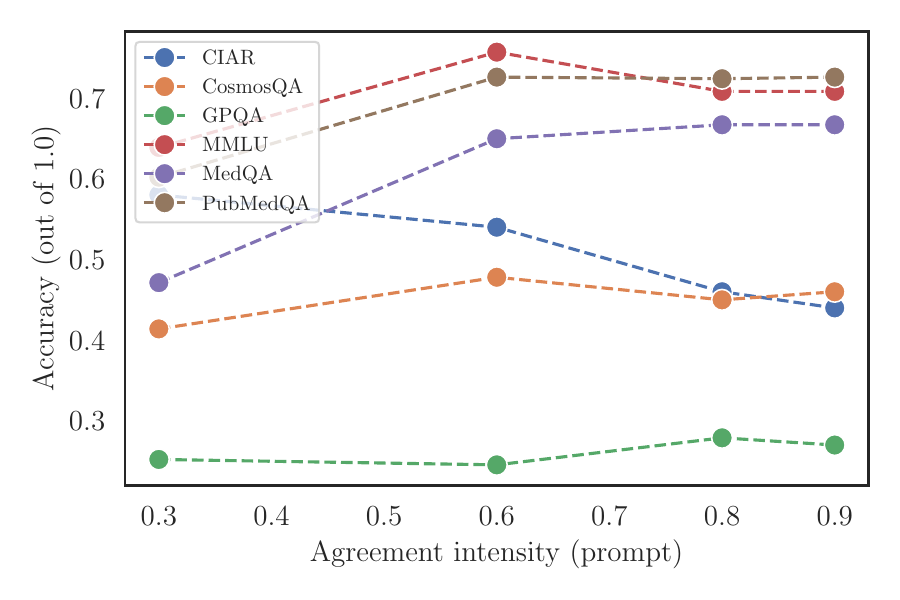}
    \includegraphics[width=0.45\textwidth]{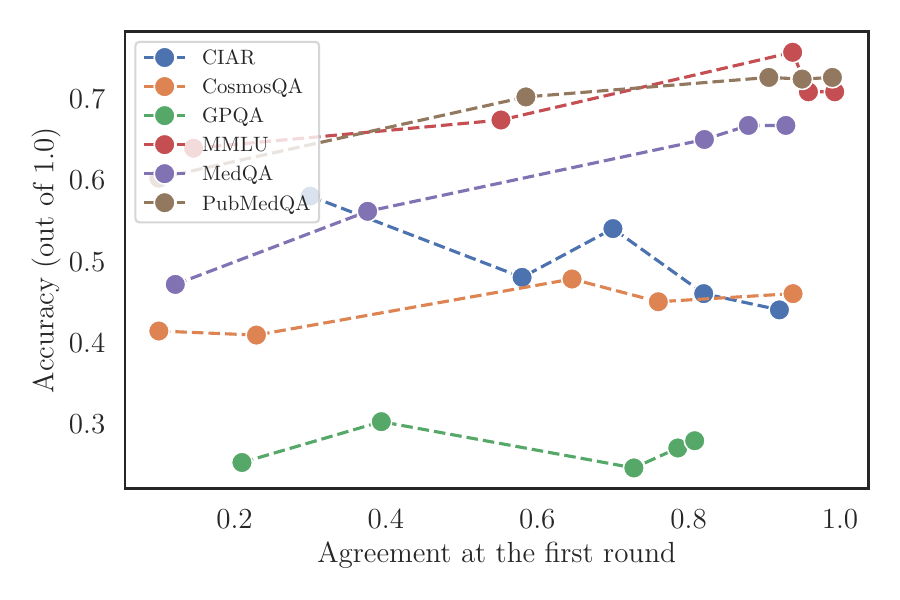}
    \caption{\textbf{Left}: Multi-Persona's accuracy according to the agreement fixed by the prompt.
    \textbf{Right}: Multi-Persona's accuracy according to the averaged agreement during the first debate round.}
    \label{fig:agreement_dataset}
\end{figure*}

\begin{figure*}[t!]
  \centering
    \includegraphics[width=0.45\textwidth]{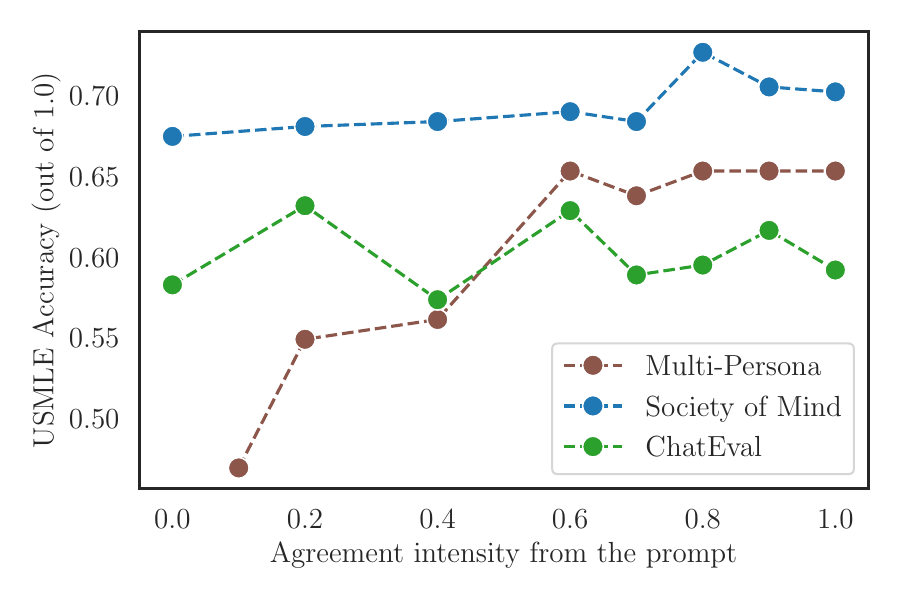}
    \includegraphics[width=0.45\textwidth]{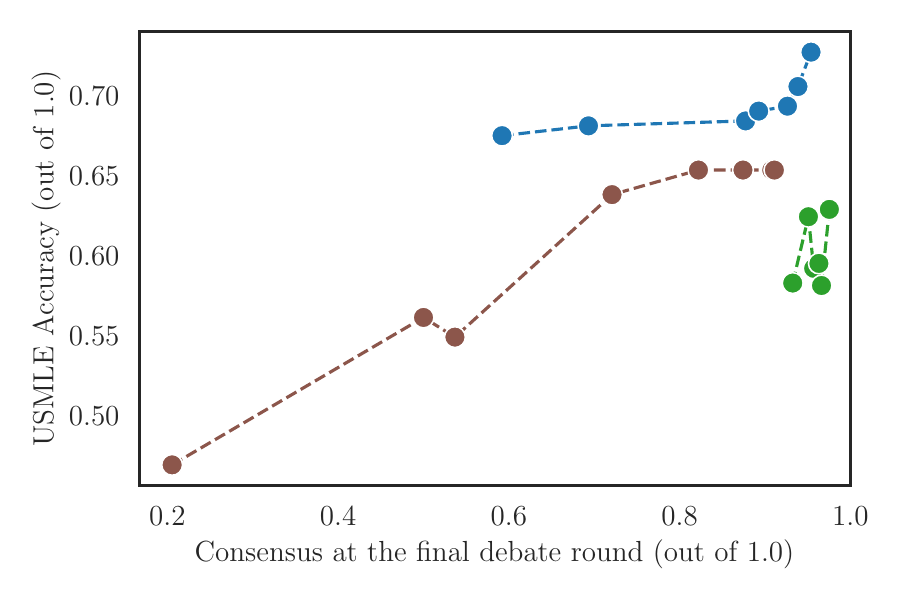}
    \caption{\textbf{Left}: Accuracy on USMLE as we increase agreement intensity in our prompt.
    \textbf{Right}: Accuracy vs actual induced debate agreement.}
    \label{fig:improvements}
\end{figure*}

An unexpected aspect of this analysis is that the Multi-Persona system reduces the overall performance compared to relying solely on the initial response of the first agent. This can be attributed to the role of the second agent (the ``devil''), which is deliberately designed to contradict or disagree, even if the initial response was correct. We show in the next section that this prior willingness for different agents to agree with each other at the outset of a debate, is indeed a very important ``hyperparameter". 

\paragraph{Improving MAD via agreement modulation}\label{sec:agreement_modulation}
Our previous experiments led us to hypothesise that the degree to which agents agree with one another during a debate may significantly affect debating performance. 
Here we test this hypothesis by developing a new MAD prompting strategy that modulates (via prompts) the degree to which agents within a debate agree with each other at the outset. Our prompt takes the following simple form: ``\textit{you should agree with the other agents X\% of the time}''. We call X in this prompt the agent's \textit{agreement intensity}. 

Multi-Persona is an ideal testbed for this method, as it uses only two agents with different system prompts. Specifically, we provide the ``angel'' agent with the question, and we modulate the disagreement using the ``devil'''s system prompt. Figure~\ref{fig:agreement_dataset} shows the effect of modulating the agent's agreement on the 6 datasets. We can see that this parameter has a significant effect on the performance and that modulating the agreement in the prompt has the effect of modifying the agreement on the first debate round, which in turn affects the outcome and the final accuracy. Strikingly, the tendency is not always the same for each dataset: while MedQA and PubMedQA directly benefit from a high agreement, CIAR follows a reverse pattern. As CIAR was created to be counter-intuitive, it is therefore probable that prompting the second agent to strongly disagree raises the chance of arriving at the correct answer during the debate. 

To investigate further, we select the highest performing configurations of debating approaches from Figure~\ref{fig:medqa_scatter}, namely SoM, ChatEval and Multi-Persona, and a subset of MedQA dataset (376 multi-choice USMLE Q\&A~\citep{han2023medalpaca}). We opted for this subset due to its high-quality questions.
%
In Figure~\ref{fig:improvements} (left), we plot the performance of each strategy as we increase the prompted agreement intensity from zero to 100\%. 
Figure~\ref{fig:improvements} (right) we plot the accuracy vs the actual observed debate consensus, i.e.\ how frequently all the agents agree upon a final answer at the end of the debate. 

We can see that modulating the agreement intensity in this way provides a substantial ($\approx$15\%) improvement in performance for Multi-Persona, and ($\approx$5\%) for SoM on the USMLE dataset. ChatEval, on the contrary, is hardly affected by this prompting mechanism, suggesting that the agent's agreement is not as easily modulable for every system.
Building on this finding, we apply the 90\% agreement intensity agent prompts to Multi-Persona on the full MedQA dataset and demonstrate (as far as we are aware) a new state-of-the-art result (for GPT-3), highlighted in Figure~\ref{fig:medqa_scatter} by the red cross symbol. 



\begin{figure*}[hbt!]
  \centering
   \begin{subfigure}[b]{0.55\textwidth}
    \includegraphics[width=\textwidth]{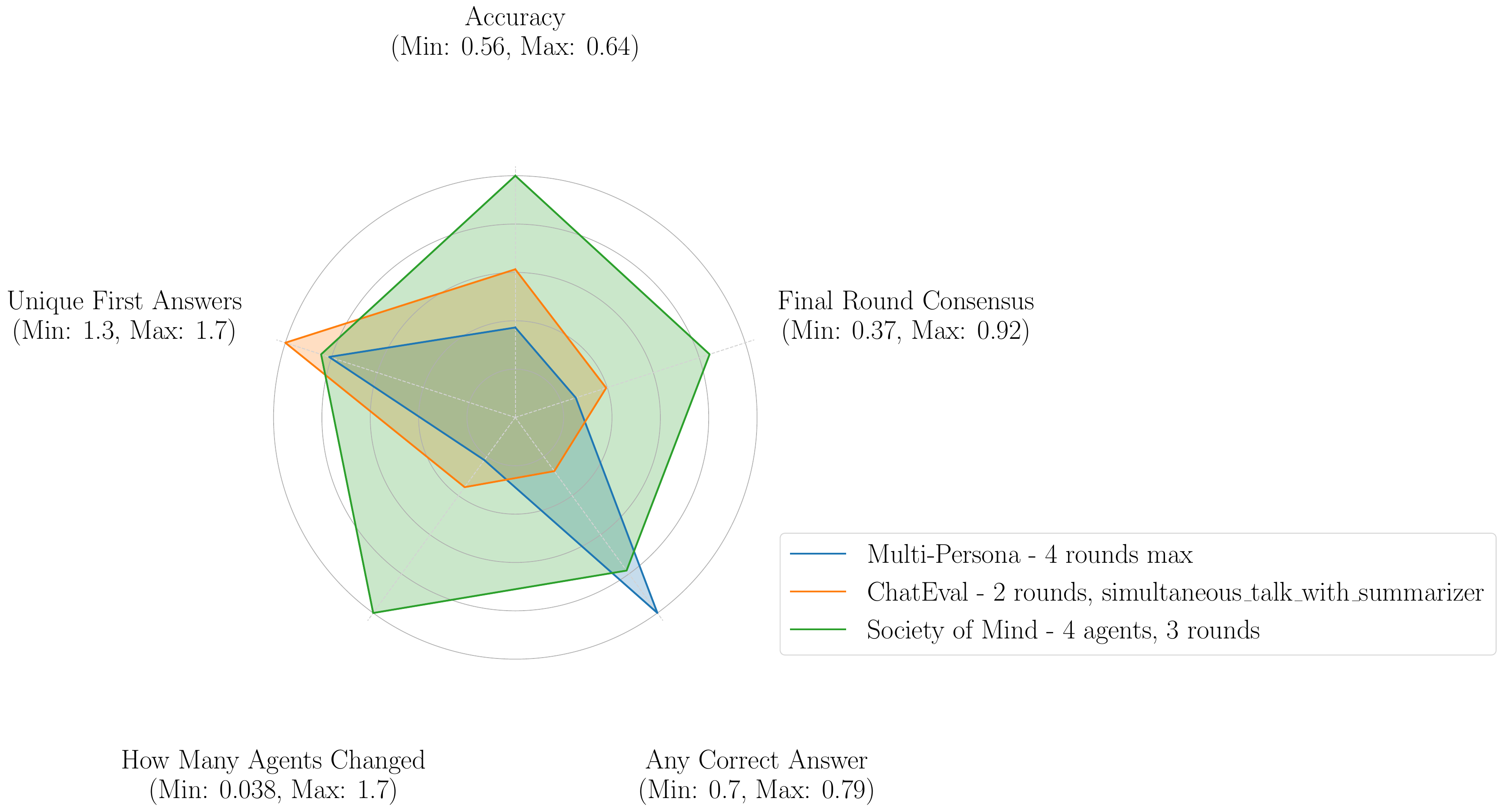}
    \caption{Best system runs}
  \end{subfigure}
   \begin{subfigure}[b]{0.42\textwidth}
    \includegraphics[width=\textwidth]{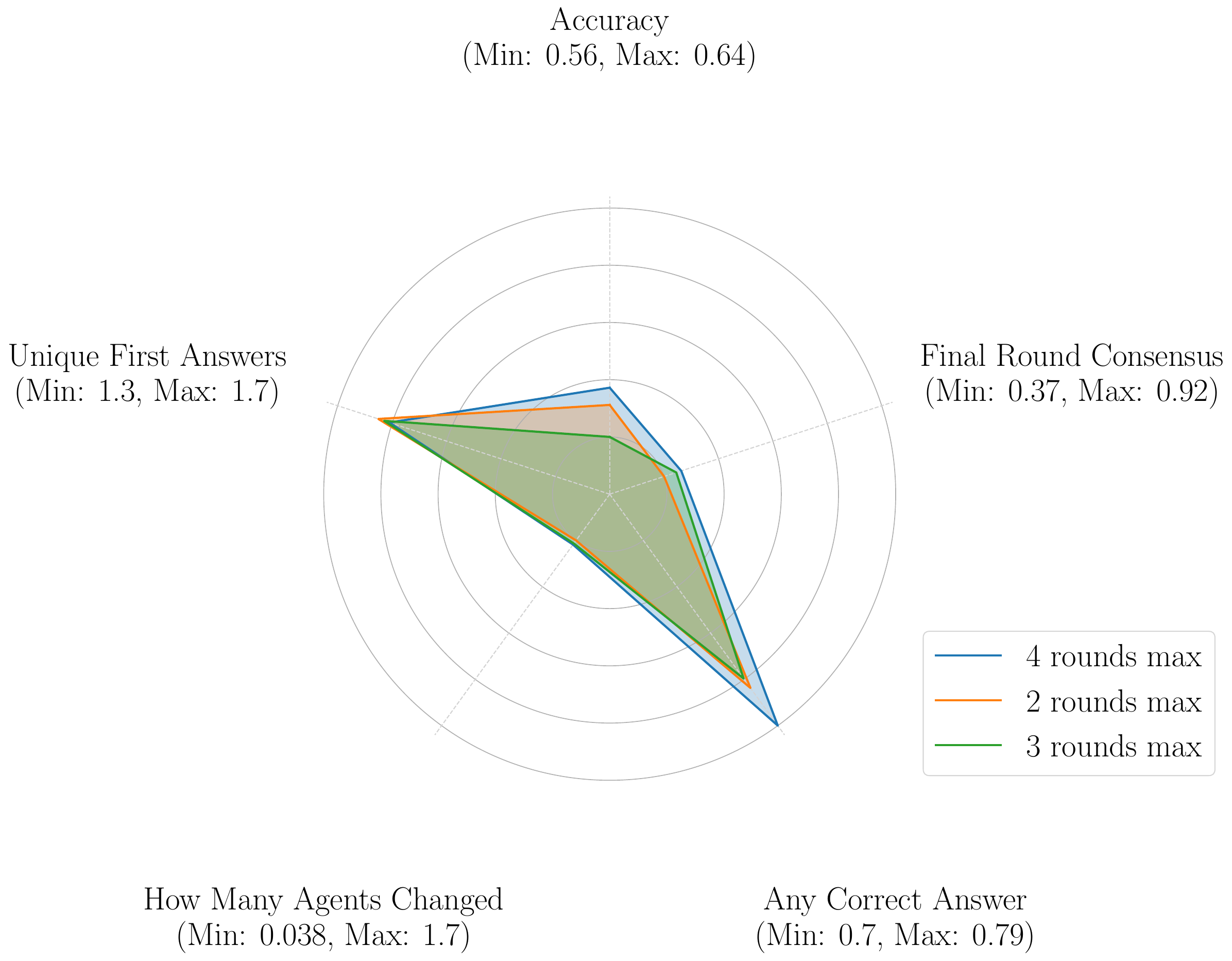}
    \caption{Multi-Persona}
  \end{subfigure}
  \begin{subfigure}[b]{0.54\textwidth}
    \includegraphics[width=\textwidth]{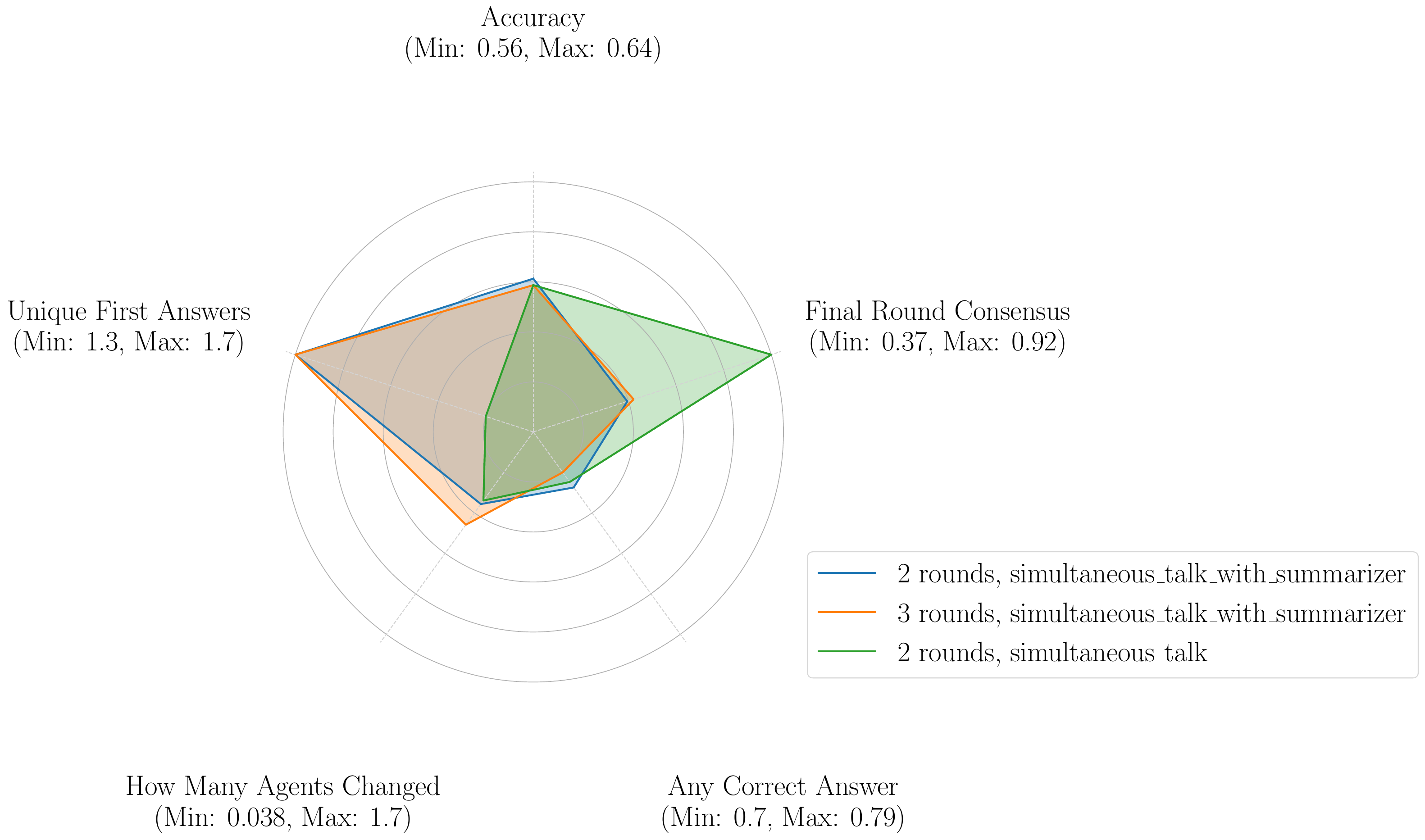}
    \caption{ChatEval}
  \end{subfigure}
  \begin{subfigure}[b]{0.44\textwidth}
    \includegraphics[width=\textwidth]{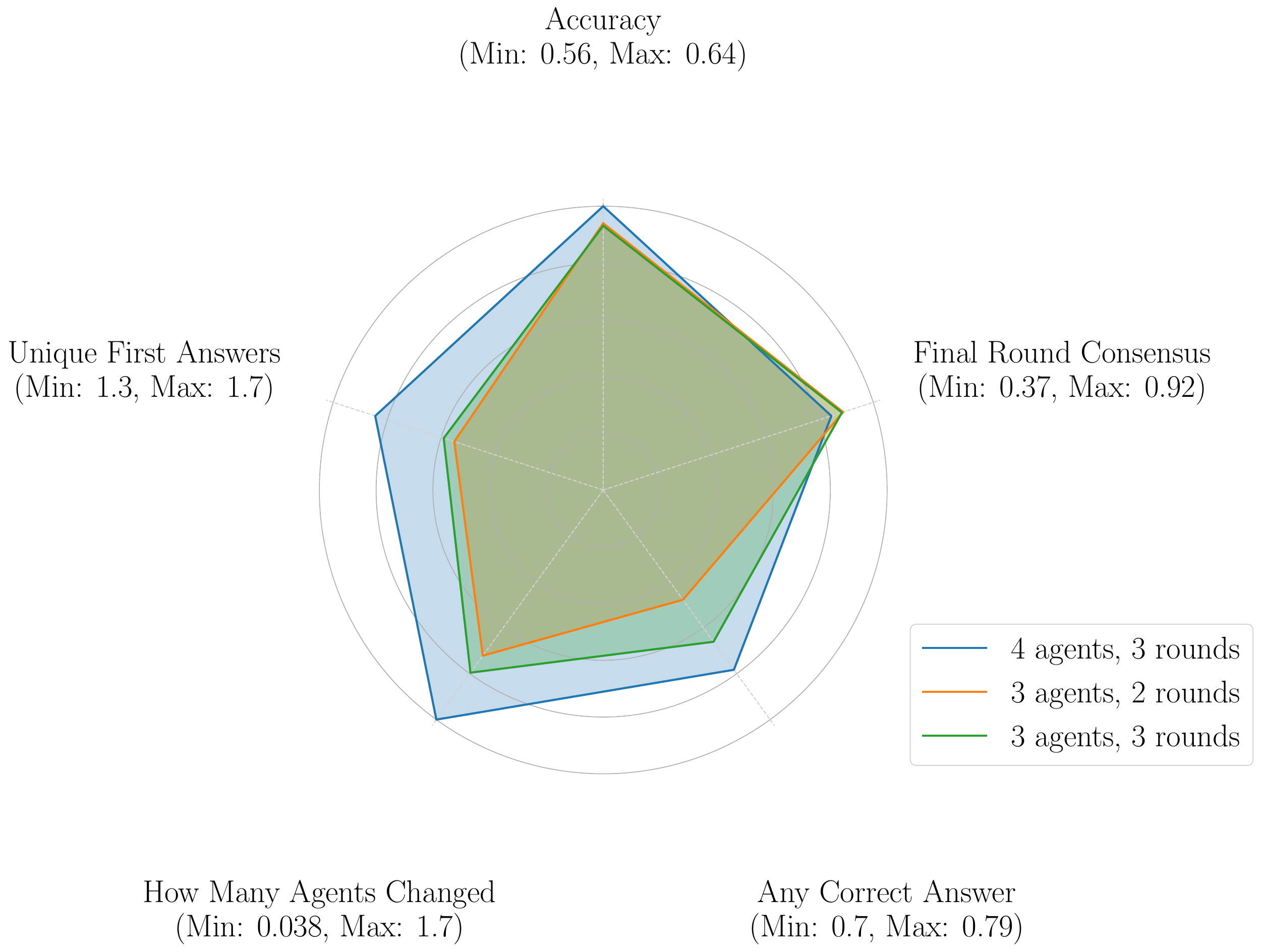}
    \caption{Society of Mind}
  \end{subfigure}
  
  \hfill

  \hfill
  \caption{\emph{Debating Behavior of MAD Strategies.} The first plot compares the top runs for each of the systems, while the other three plots compare the top three runs for each system. Each plot quantifies the performance on MedQA by measuring several factors: the accuracy of the final consensus among agents (scaled to 1.0), the instances of any agent identifying the correct answer independently (also scaled to 1.0), the frequency of agents changing their answers during the debate, and the diversity of initial answers provided. The descriptions for each axis include the range of scores observed, from minimum to maximum.}
  \label{fig:debate_metrics}
\end{figure*}

\paragraph{Analysis of debating behaviour} \label{app:additional-metrics}
Our experiments section concludes with an exploration of the dynamics within MAD systems, specifically through their performance on the MedQA test set. We assess Multi-Persona, ChatEval, and Society of Mind, focusing on key aspects such as consensus accuracy, independent correct answer identification, the adaptability reflected in answer changes, and the breadth of initial perspectives. These facets are depicted in Figure \ref{fig:debate_metrics}, offering insights into each system's strategic approach to debate and problem-solving.


The four spider plots offer a more detailed analysis of the debating behavior of MAD strategies, providing insights into the effectiveness of each system through various metrics. The initial plot compares the three systems, with Society of Mind (SoM) notably exhibiting the largest surface area, suggesting it outperforms others across all debating metrics. This is further supported in the subsequent plots, which analyze the top three runs for each system and show that larger areas are indicative of better performance. These visualizations emphasize the importance of a balanced system, indicating that excellence in one aspect is not enough; a system must perform well across multiple dimensions of the debate process. A full description of each of the metrics is provided in Appendix \ref{app:all_debate_metrics}.

For example, when agents change their answers during a debate, this reevaluation must lead to more accurate conclusions. Arbitrary changes without a strategic basis could prevent consensus. This underlines the need for a balanced approach where agents not only reconsider their positions but also enhance their decision-making accuracy, contributing to the system's overall effectiveness.

\begin{figure*}[t!]
  \centering
    {\includegraphics[width=0.54\textwidth]{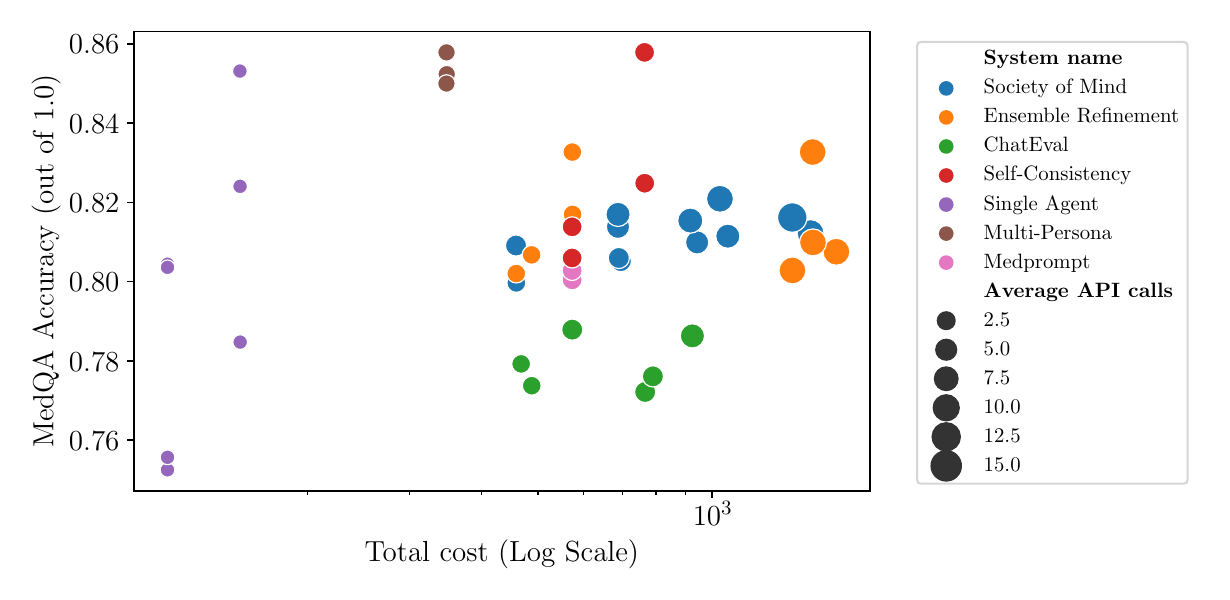}}
    {\includegraphics[width=0.41\textwidth]{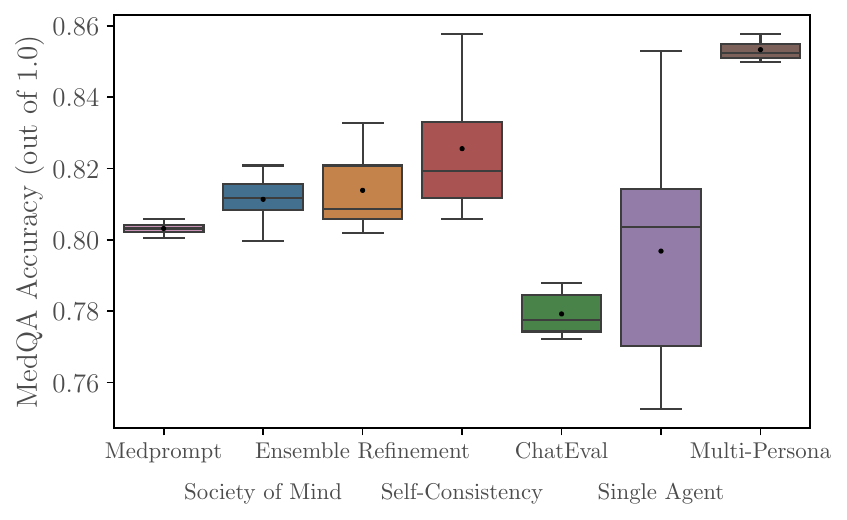}}
    \caption{\textit{Benchmark of experiment configurations of GPT-4 on MedQA dataset.} \textbf{Left}: Accuracy vs average cost (\$) per question. The size of the dots reflects the average number of API calls required per question. 
    \textbf{Right}: Summarizes accuracy grouped by strategy, sorted by average performance (black dot).} 
    \label{fig:gpt4_medqa_scatter}
\end{figure*}

\begin{figure*}[t!]
  \centering
    {\includegraphics[width=0.54\textwidth]{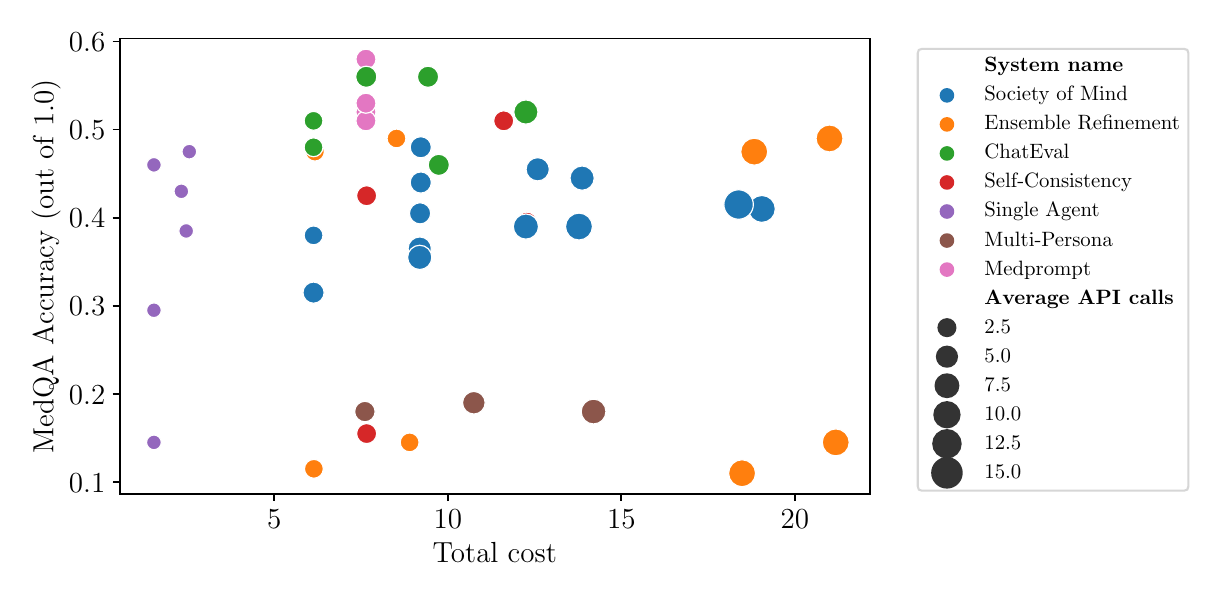}}
    {\includegraphics[width=0.41\textwidth]{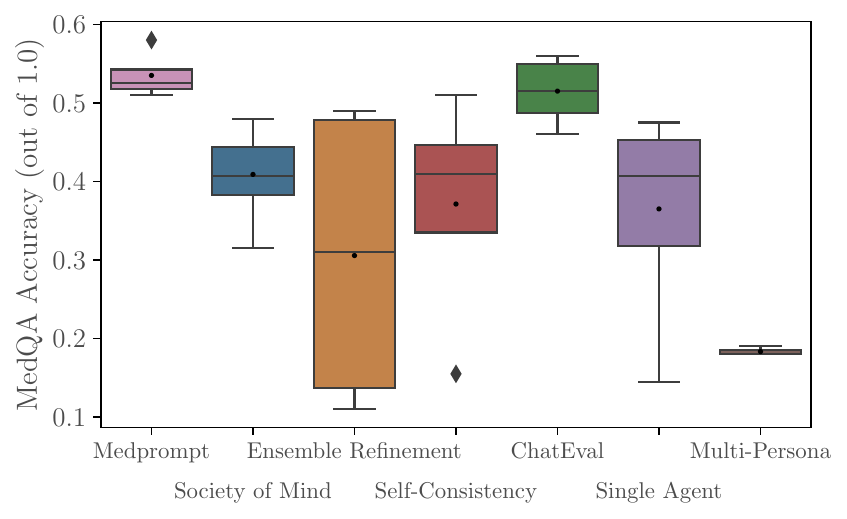}}
    \caption{\textit{Benchmark of experiment configurations of Mixtral 8x7B on MedQA dataset.} \textbf{Left}: Accuracy vs average cost (\$) per question. The size of the dots reflects the average number of API calls required per question. 
    \textbf{Right}: Summarizes accuracy grouped by strategy, sorted by average performance (black dot).} 
    \label{fig:mixtral_medqa_scatter}
\end{figure*}

\paragraph{Evaluating using other APIs}
We assess the capability of GPT-4 \citep{openai2024gpt4} and Mixtral 8x7B \citep{jiang2024mixtral} on the MedQA dataset, applying the optimal agreement modulation value identified for Multi-Persona with GPT-3.5 on USMLE. Results shown in Figures~\ref{fig:gpt4_medqa_scatter} and~\ref{fig:mixtral_medqa_scatter}.

The results from GPT-3.5 suggest that hyperparameter settings can be effectively transferred to GPT-4. However, this transferability does not extend well to Mixtral 8x7B. This discrepancy might be attributed to architectural differences, which could affect how hyperparameters influence model performance. We leave this for future research.

\paragraph{Code availability} Source code for this work, including all MAD implementations and configurations, is publicly available at \url{https://github.com/instadeepai/DebateLLM}.

\section{Conclusions} \label{sec:conclusion}
The investigations detailed in this work have demonstrated that MAD approaches currently do not outperform other ensembling methods such as Medprompt and self-consistency using their original implementations. 
MAD typically requires a higher number of API calls, increasing the number of tokens to produce and process and ultimately, the total running cost of the system. Interestingly, we found that this might primarily be due to MAD protocols being more sensitive to hyperparameters and not necessarily that debate inherently performs worse than single-agent methods.
To strengthen this claim, we demonstrated that a simple prompt-based manipulation of the degree to which agents agree with each other, referred to as the agreement intensity, could provide significant performance gains. By tuning this agreement intensity for Multi-Persona, it went from being the worst-performing protocol to the best-performing, beating out methods such as Medprompt and Self-Consistency.  
Therefore, while MAD systems currently trail behind other prompting strategies, we believe they hold significant potential to transform the prompting landscape. To foster further exploration, we have made our code repository open-source, for others to easily reproduce all our experiments and further pursue interesting related research directions.
\paragraph{Limitations} We utilize API calls to a publicly available LLM~\citep{brown2020language} which, whilst sufficient in the context of our investigation, exposes us to variable inference time calls and unforeseen model updates. Moreover, large-scale API-based benchmarking incurs substantial financial and time costs, which both limit the experiment scales and provide a barrier of entry to replication and extension efforts.  For these reasons, future works could extend this line of work using open-source models and in-house infrastructure.

\section*{Impact Statement} \label{sec:impact_statement}
This study furthers the field of machine learning by evaluating multi-agent debate (MAD) strategies and other prompting methods within question-answering datasets. Our results show that MAD approaches could be as effective as, or even superior to, other strategies like Medprompt and Self-Consistency. An open-source toolkit is provided to promote additional research in this domain. While this research aims to improve the precision of large language models (LLMs) in answering questions, with a significant focus on healthcare applications, it is crucial to consider the potential risks associated with these advancements. Specifically, medical question-answering systems can sometimes yield incorrect predictions in unforeseen ways, and there is a risk that these systems may exhibit undue confidence in their erroneous responses. Such inaccuracies, especially in the medical domain, could lead to misinformation or misdiagnosis, potentially impacting patient care and outcomes. Therefore, while this work contributes valuable insights and tools for enhancing LLM capabilities, it is imperative to proceed with caution, incorporating robust validation, transparency, and ethical considerations to mitigate risks and ensure these technologies are deployed in a manner that safeguards public health and trust.

\small

\bibliography{icml2024}
\bibliographystyle{icml2024}

\newpage
\appendix
\onecolumn
\section{Appendix}


  
  
  

\subsection{Extended results on additional datasets} \label{app:results_overview}
\begin{figure}[h!]
  \centering
  \begin{subfigure}[h]{0.49\linewidth}
    \centering
    \includegraphics[height=3.9cm, keepaspectratio]{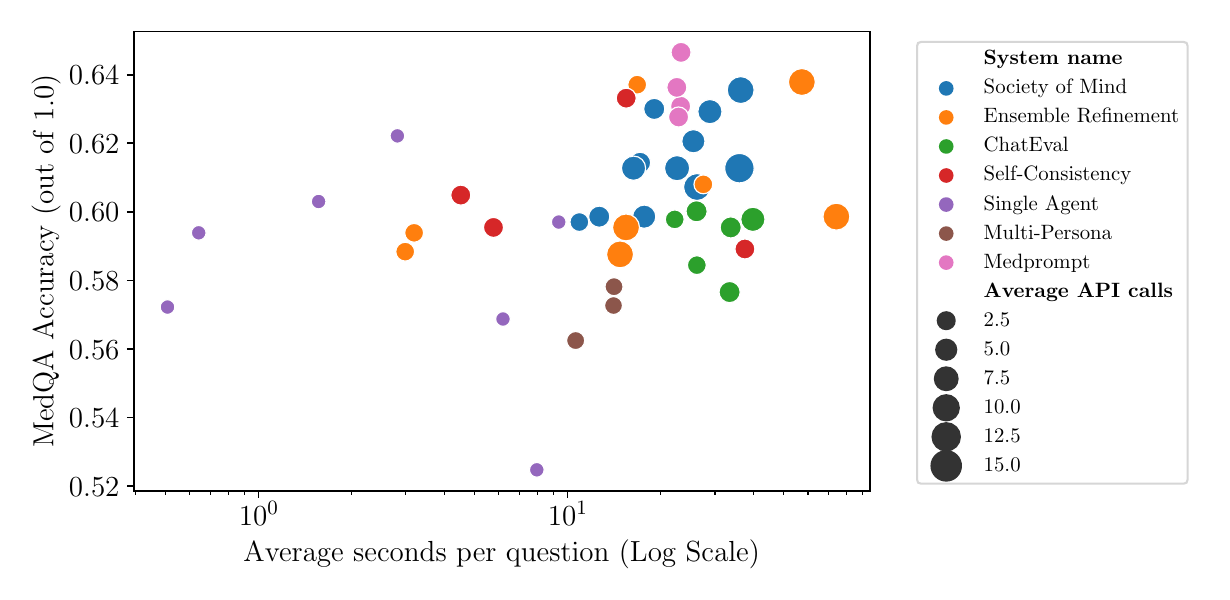}
    \caption{Accuracy versus the average time per question}
  \end{subfigure}
  \hfill
  \begin{subfigure}[h]{0.49\textwidth}
    \centering
    \includegraphics[height=3.9cm, keepaspectratio]{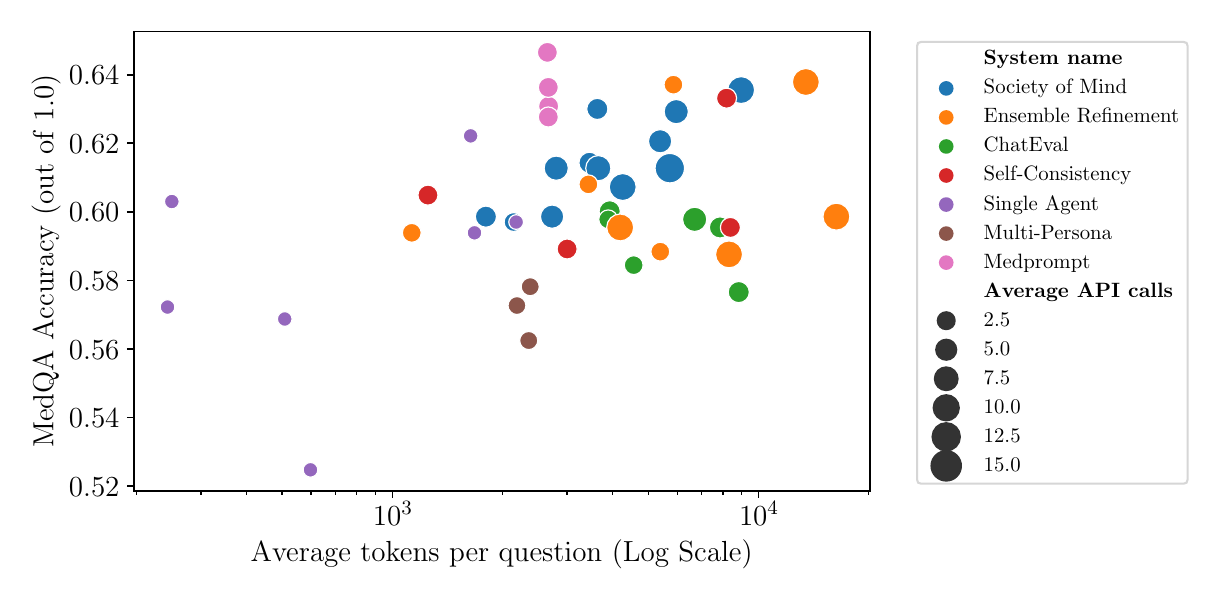}
    \caption{Accuracy versus Average tokens per question}
  \end{subfigure}
  
  \begin{subfigure}[h]{0.49\textwidth}
    \centering
    \includegraphics[width=\textwidth]{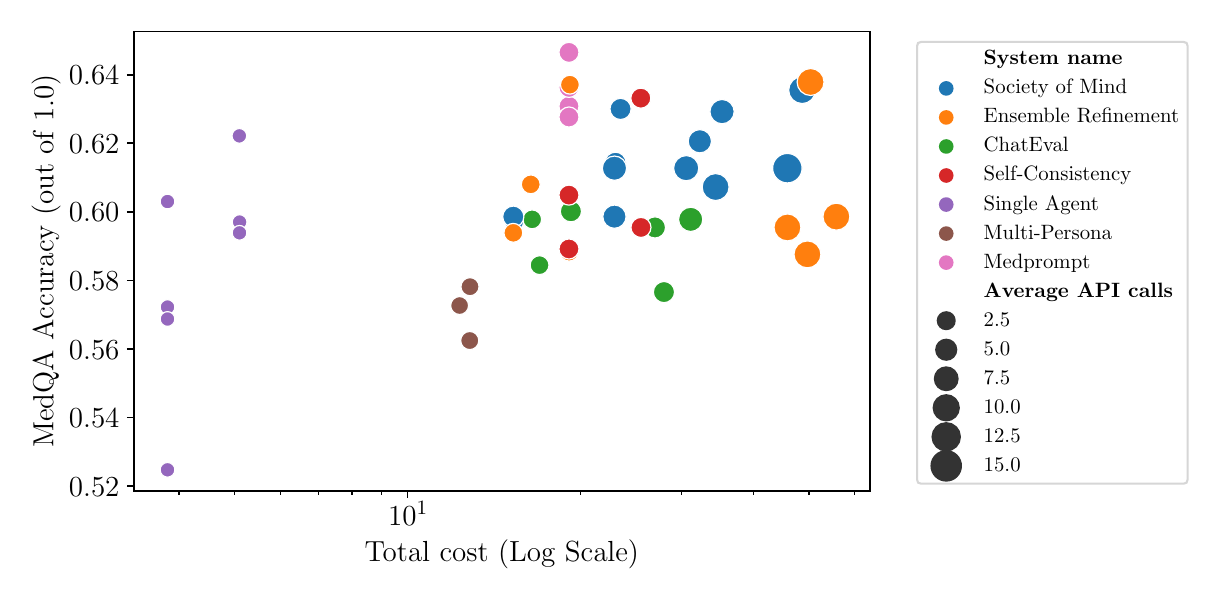}
    \caption{Accuracy versus the total cost}
  \end{subfigure}
  \hfill 
  \begin{subfigure}[h]{0.38\textwidth}
    \centering
    \hspace*{-40mm} 
    \includegraphics[width=\textwidth]{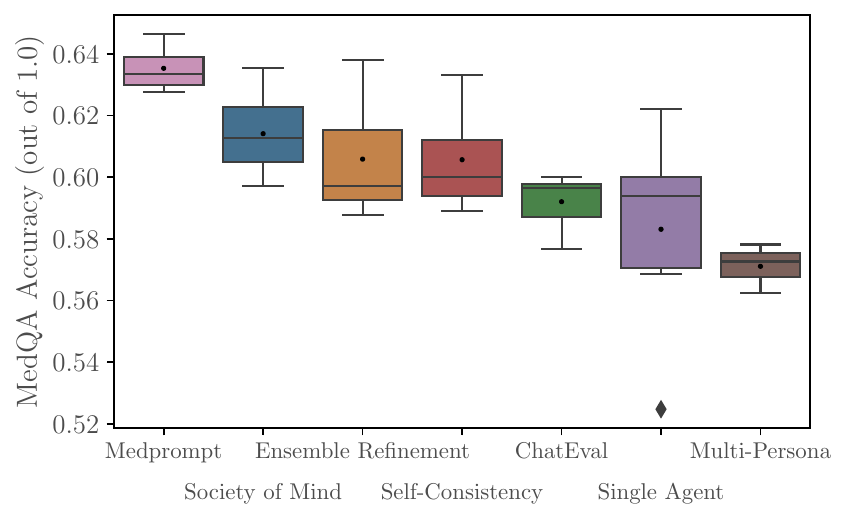}
    \caption{Accuracy by strategy}
  \end{subfigure}
\caption{MedQA experimental results.}
\label{fig:medqa_results}
\end{figure}

\begin{figure}[H]
  \centering
  \begin{subfigure}[h]{0.49\linewidth}
    \centering
    \includegraphics[height=3.9cm, keepaspectratio]{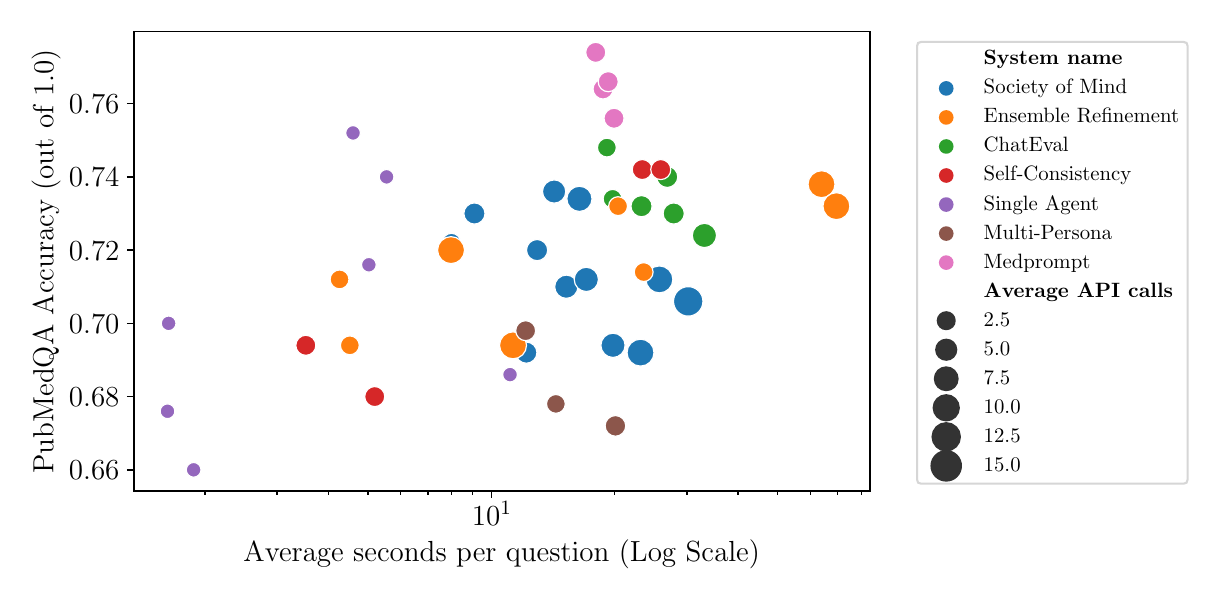}
    \caption{Accuracy versus the average time per question}
  \end{subfigure}
  \hfill
  \begin{subfigure}[h]{0.49\textwidth}
    \centering
    \includegraphics[height=3.9cm, keepaspectratio]{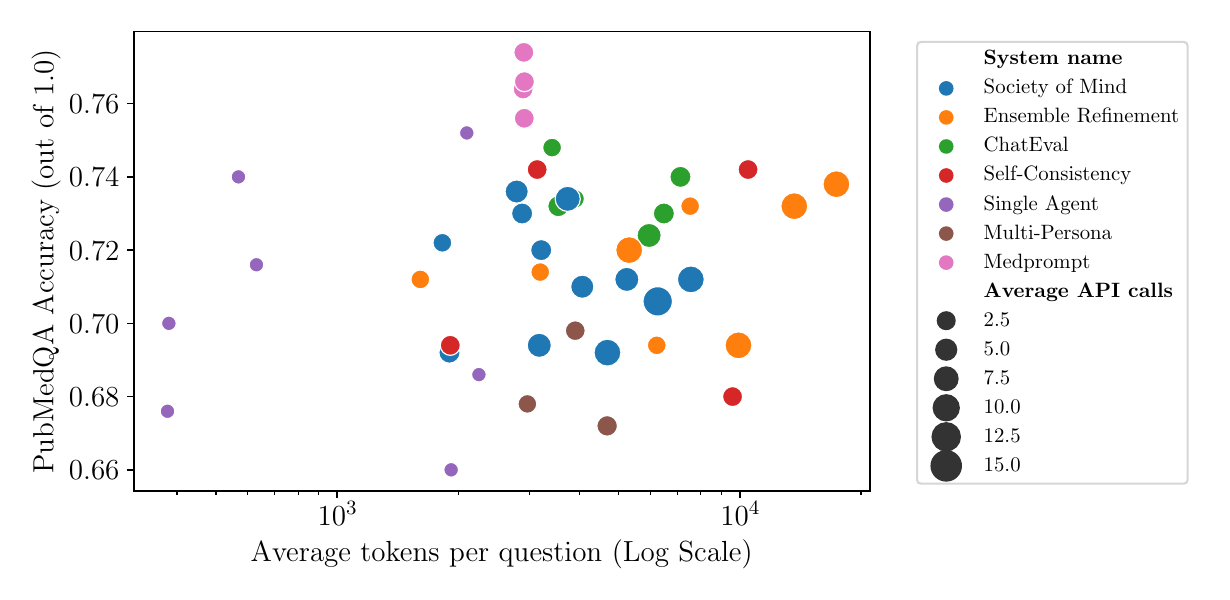}
    \caption{Accuracy versus Average tokens per question}
  \end{subfigure}
  
  \begin{subfigure}[h]{0.49\textwidth}
    \centering
    \includegraphics[width=\textwidth]{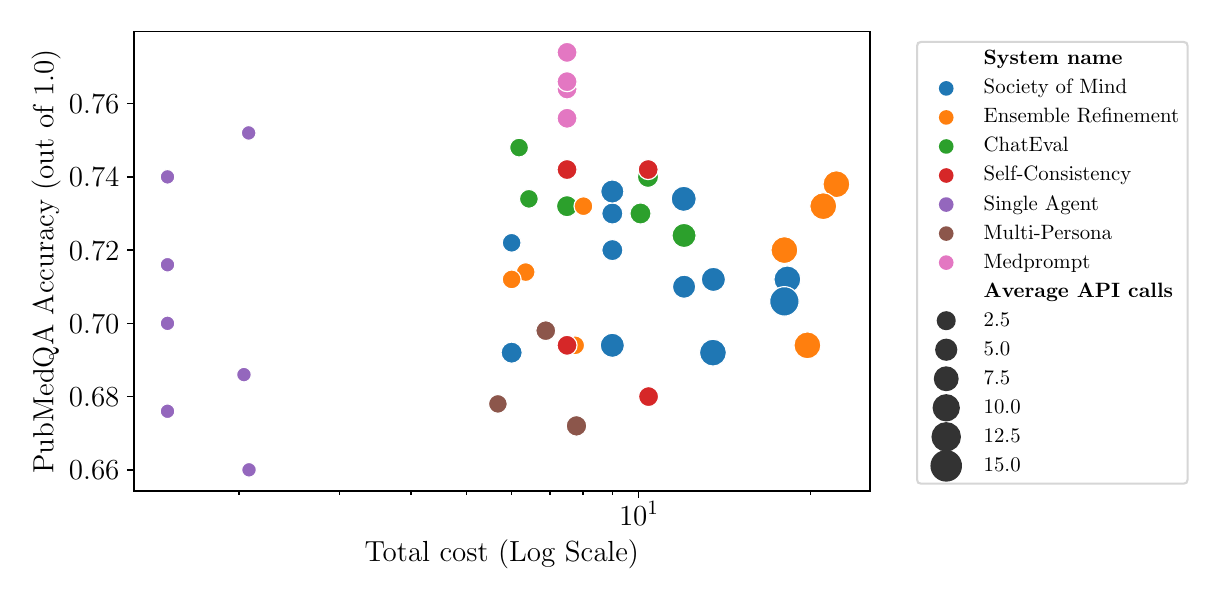}
    \caption{Accuracy versus the total cost}
  \end{subfigure}
  \hfill 
  \begin{subfigure}[h]{0.38\textwidth}
    \centering
    \hspace*{-40mm} 
    \includegraphics[width=\textwidth]{imgs/pubmedqa_total_acc_box.pdf}
    \caption{Accuracy by strategy}
  \end{subfigure}
\caption{PubMedQA experimental results.}
\label{fig:pubmedqa_results}
\end{figure}

\begin{figure}[H]
  \centering
  \begin{subfigure}[h]{0.49\linewidth}
    \centering
    \includegraphics[height=3.9cm, keepaspectratio]{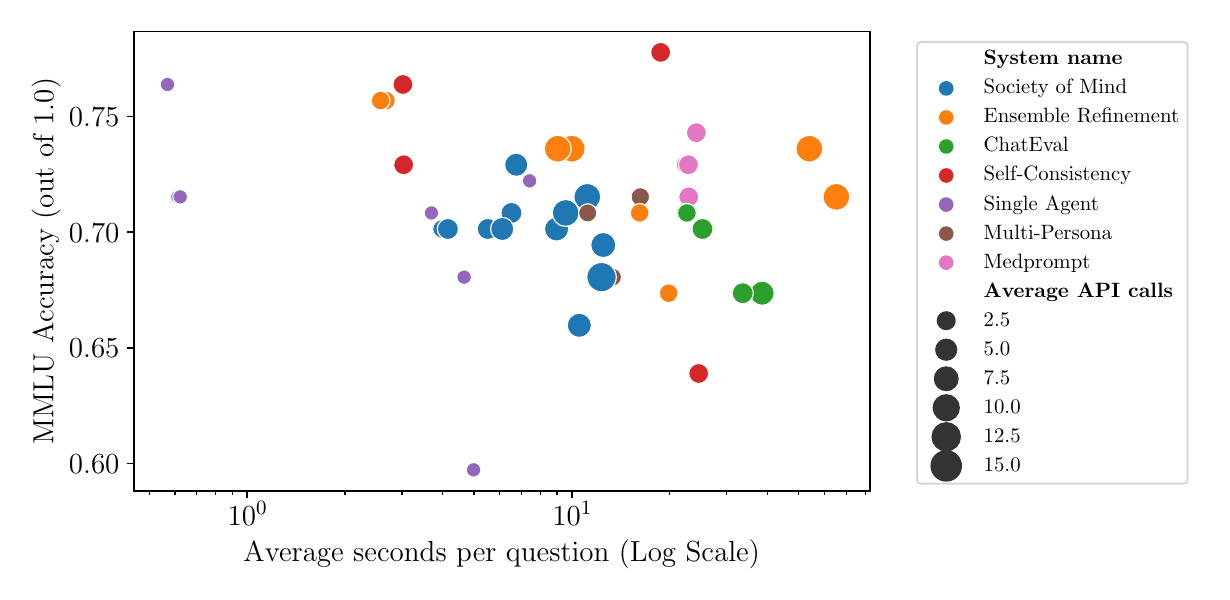}
    \caption{Accuracy versus the average time per question}
  \end{subfigure}
  \hfill
  \begin{subfigure}[h]{0.49\textwidth}
    \centering
    \includegraphics[height=3.9cm, keepaspectratio]{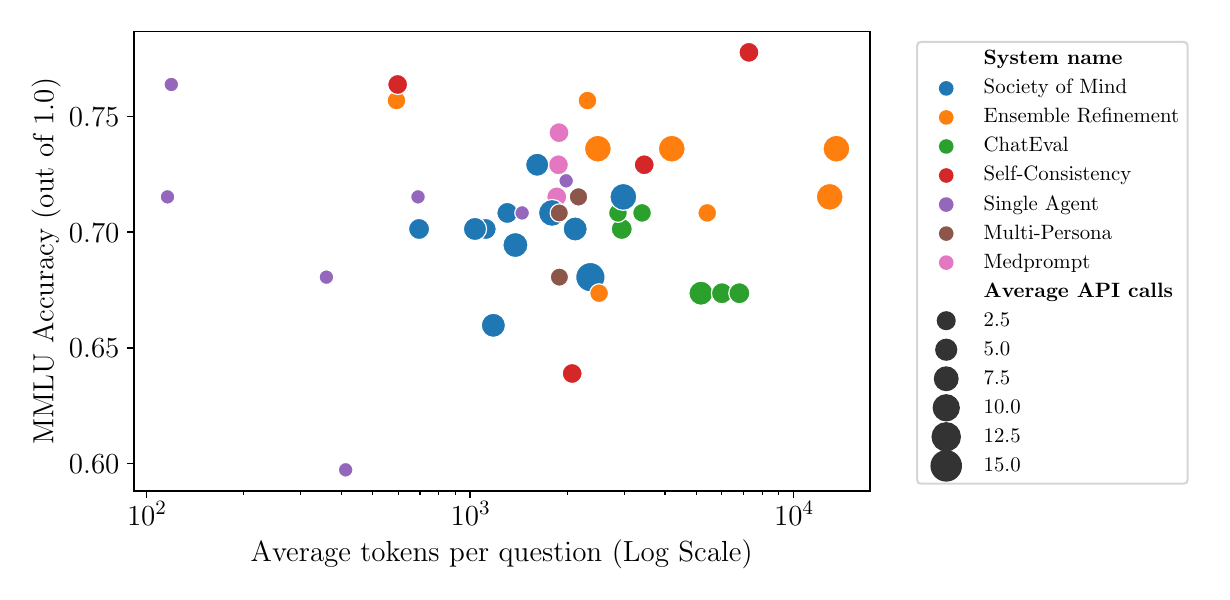}
    \caption{Accuracy versus Average tokens per question}
  \end{subfigure}
  
  \begin{subfigure}[h]{0.49\textwidth}
    \centering
    \includegraphics[width=\textwidth]{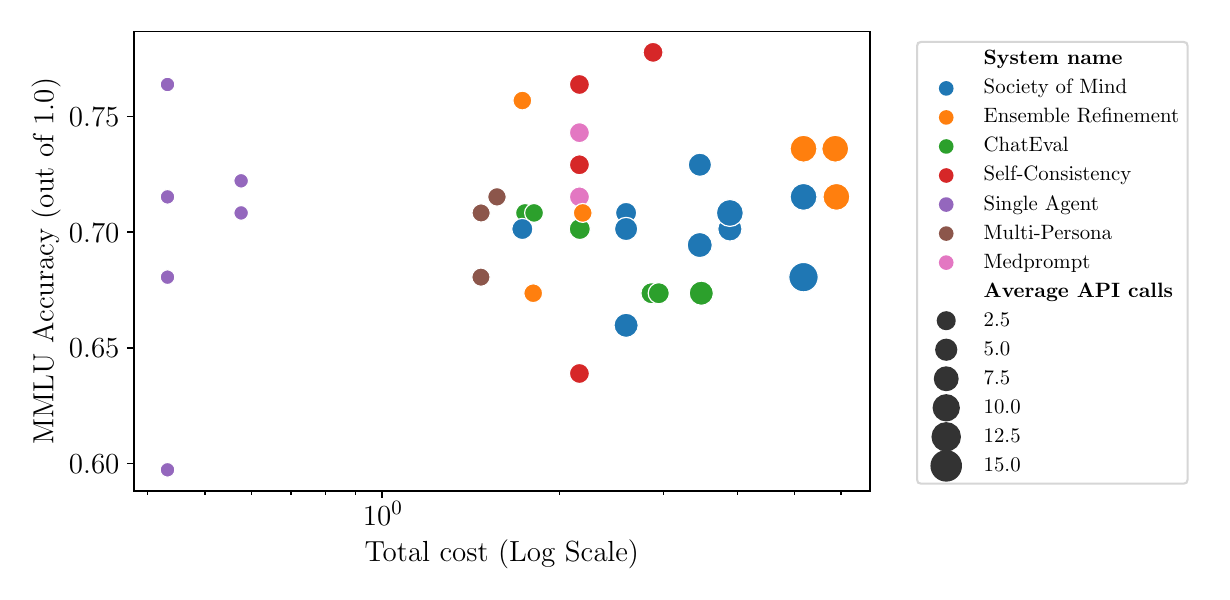}
    \caption{Accuracy versus the total cost}
  \end{subfigure}
  \hfill 
  \begin{subfigure}[h]{0.38\textwidth}
    \centering
    \hspace*{-40mm} 
    \includegraphics[width=\textwidth]{imgs/mmlu_total_acc_box.pdf}
    \caption{Accuracy by strategy}
  \end{subfigure}
\caption{MMLU experimental results.}
\label{fig:mmlu_results}
\end{figure}

\begin{figure}[H]
  \centering
  \begin{subfigure}[h]{0.49\linewidth}
    \centering
    \includegraphics[height=3.9cm, keepaspectratio]{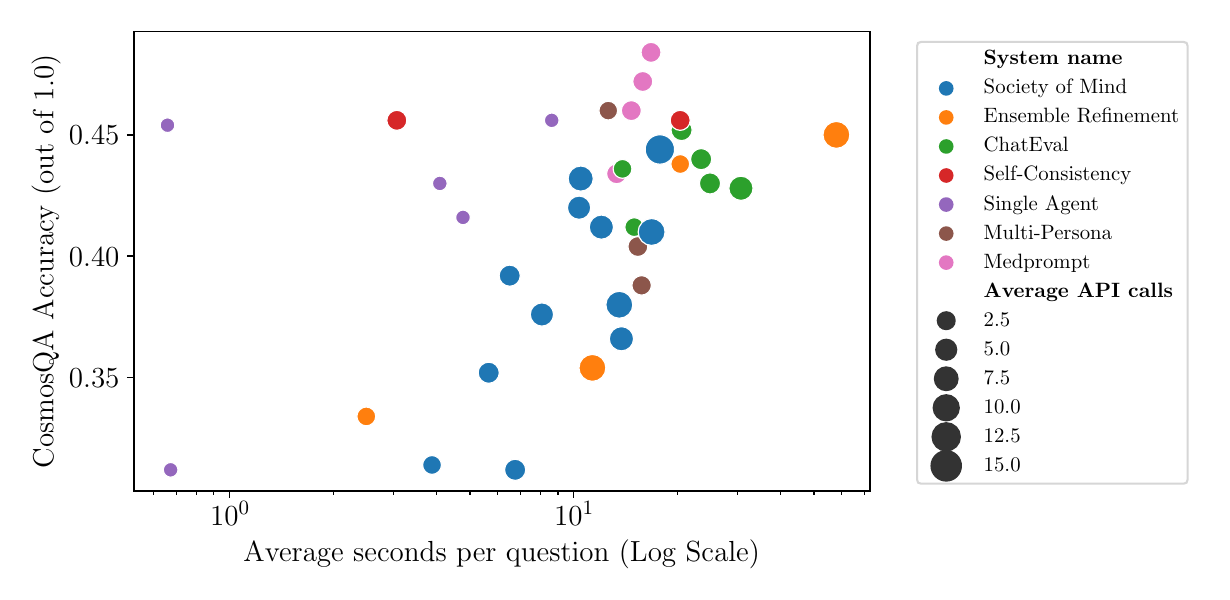}
    \caption{Accuracy versus the average time per question}
  \end{subfigure}
  \hfill
  \begin{subfigure}[h]{0.49\textwidth}
    \centering
    \includegraphics[height=3.9cm, keepaspectratio]{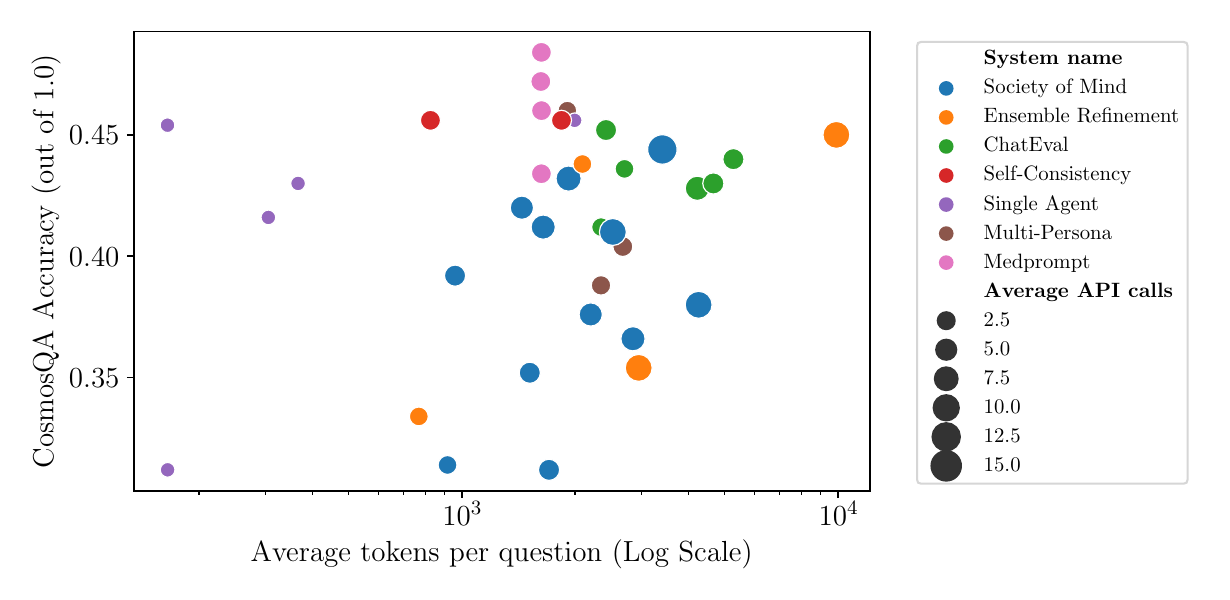}
    \caption{Accuracy versus Average tokens per question}
  \end{subfigure}
  
  \begin{subfigure}[h]{0.49\textwidth}
    \centering
    \includegraphics[width=\textwidth]{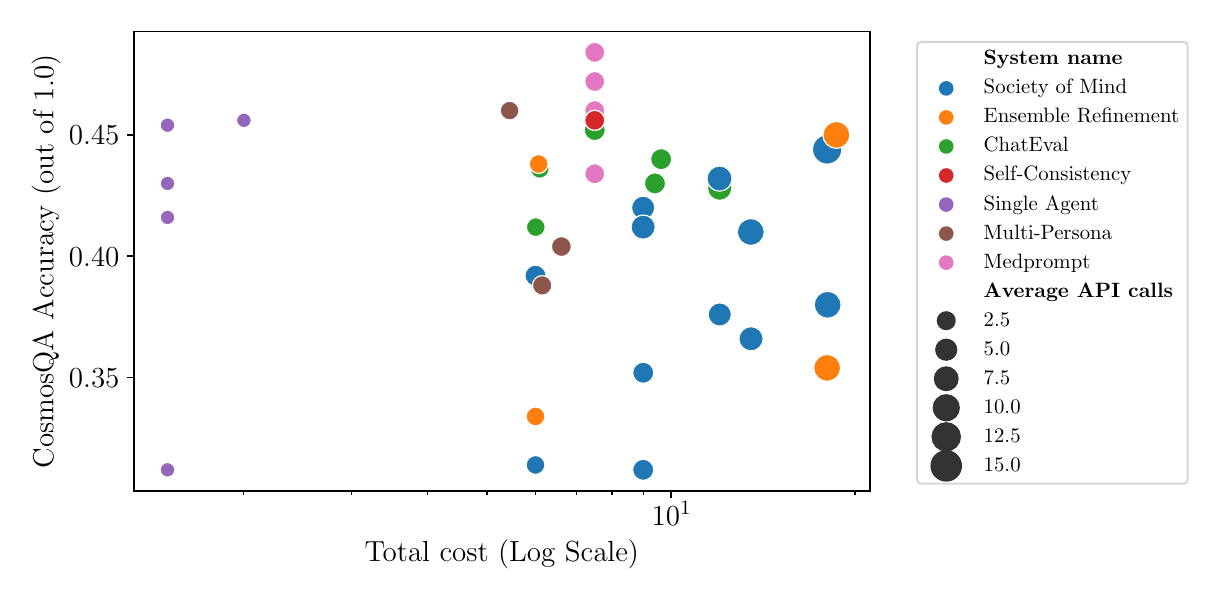}
    \caption{Accuracy versus the total cost}
  \end{subfigure}
  \hfill 
  \begin{subfigure}[h]{0.38\textwidth}
    \centering
    \hspace*{-40mm} 
    \includegraphics[width=\textwidth]{imgs/cosmosqa_total_acc_box.pdf}
    \caption{Accuracy by strategy}
  \end{subfigure}
\caption{CosmosQA experimental results.}
\label{fig:cosmosqa_results}
\end{figure}

\begin{figure}[H]
  \centering
  \begin{subfigure}[h]{0.49\linewidth}
    \centering
    \includegraphics[height=3.9cm, keepaspectratio]{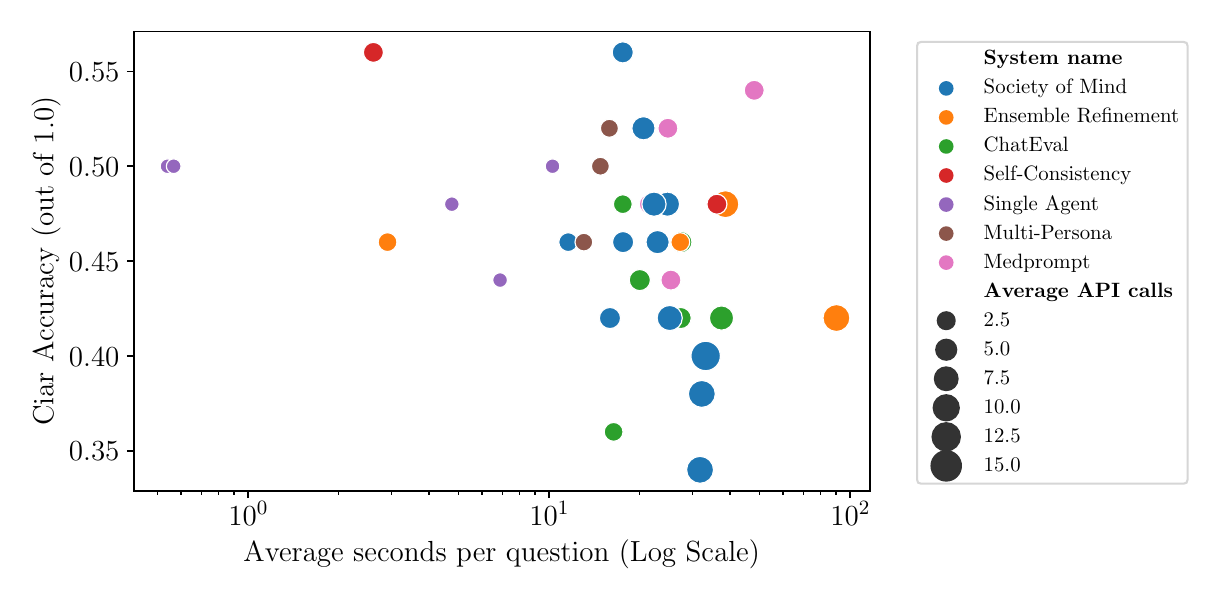}
    \caption{Accuracy versus the average time per question}
  \end{subfigure}
  \hfill
  \begin{subfigure}[h]{0.49\textwidth}
    \centering
    \includegraphics[height=3.9cm, keepaspectratio]{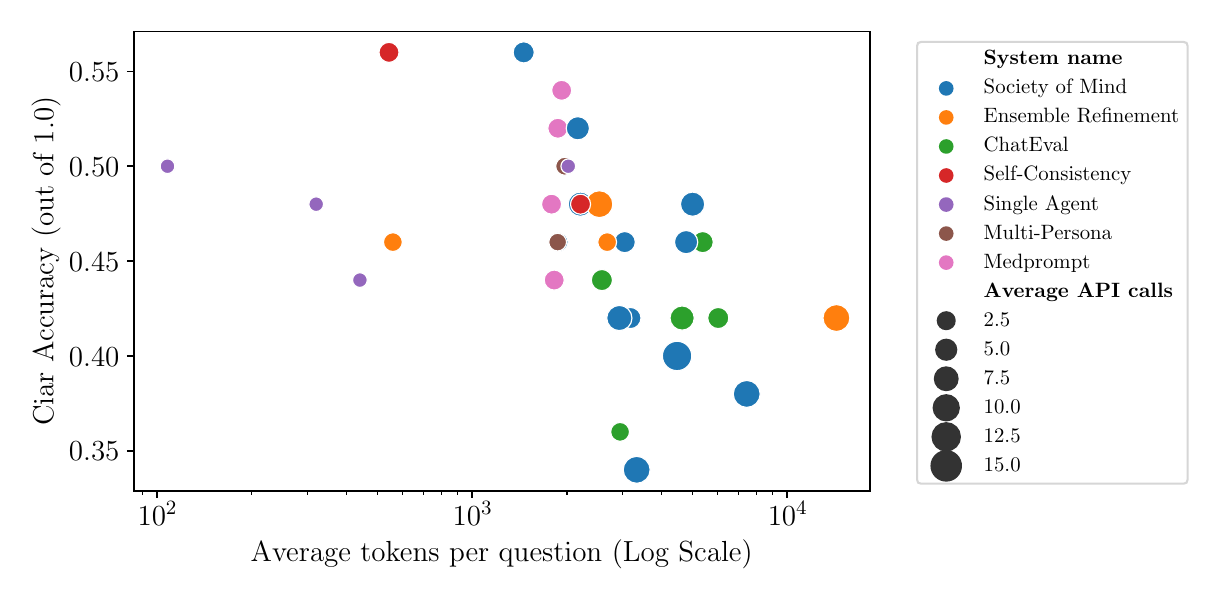}
    \caption{Accuracy versus Average tokens per question}
  \end{subfigure}
  
  \begin{subfigure}[h]{0.49\textwidth}
    \centering
    \includegraphics[width=\textwidth]{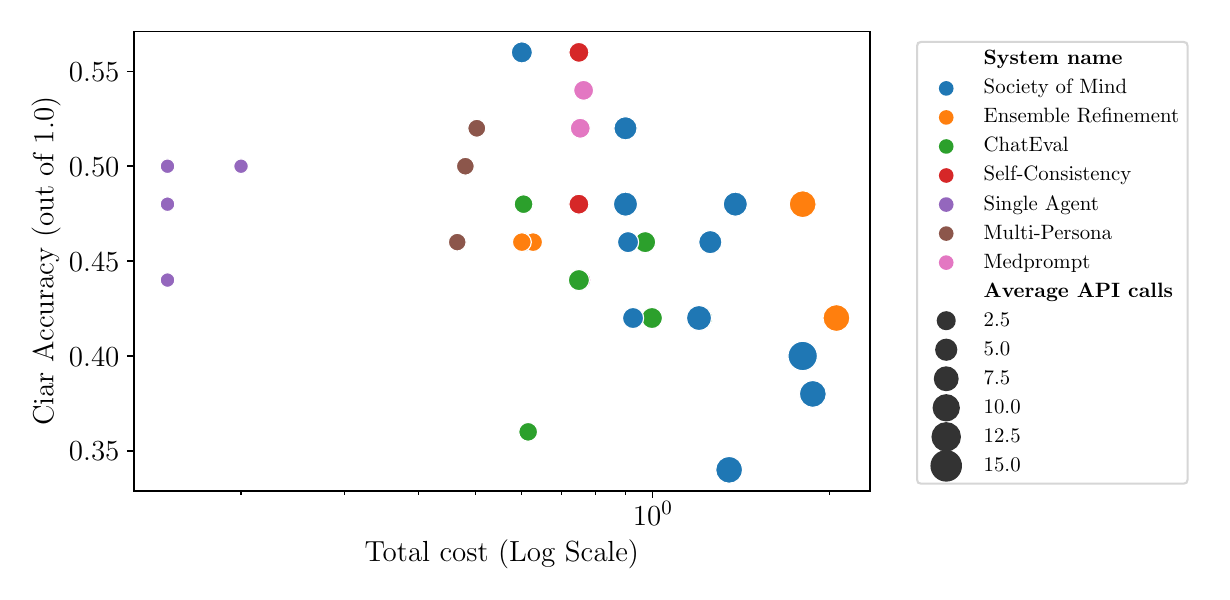}
    \caption{Accuracy versus the total cost}
  \end{subfigure}
  \hfill 
  \begin{subfigure}[h]{0.38\textwidth}
    \centering
    \hspace*{-40mm} 
    \includegraphics[width=\textwidth]{imgs/ciar_total_acc_box.pdf}
    \caption{Accuracy by strategy}
  \end{subfigure}
\caption{CIAR experimental results.}
\label{fig:ciar_results}
\end{figure}

\begin{figure}[H]
  \centering
  \begin{subfigure}[h]{0.49\linewidth}
    \centering
    \includegraphics[height=3.9cm, keepaspectratio]{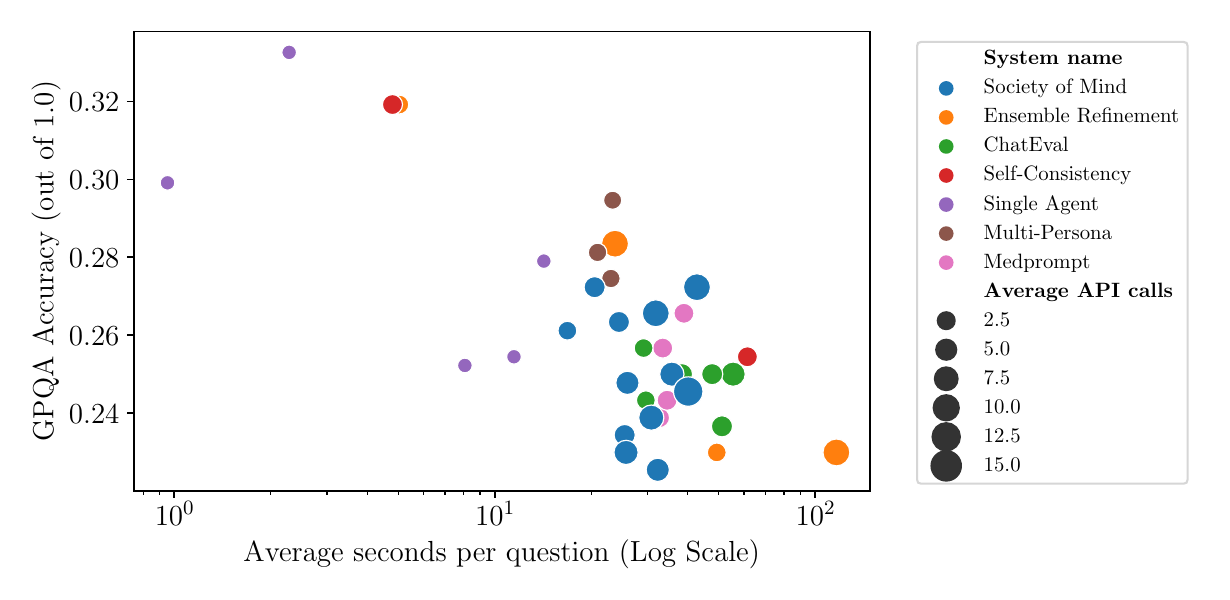}
    \caption{Accuracy versus the average time per question}
  \end{subfigure}
  \hfill
  \begin{subfigure}[h]{0.49\textwidth}
    \centering
    \includegraphics[height=3.9cm, keepaspectratio]{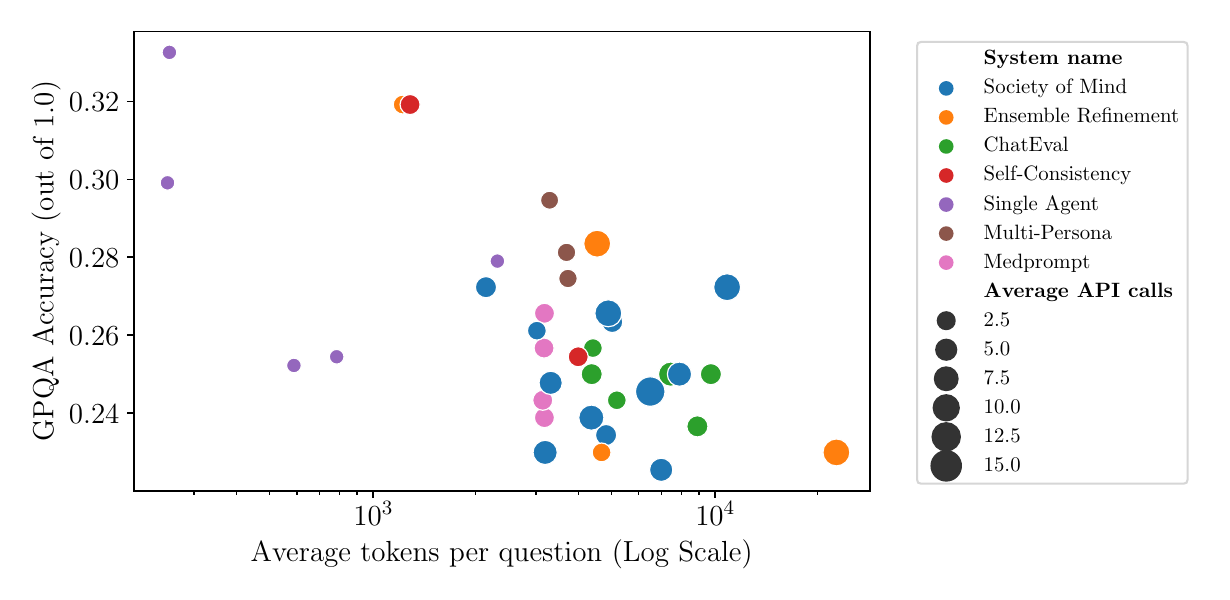}
    \caption{Accuracy versus Average tokens per question}
  \end{subfigure}
  
  \begin{subfigure}[h]{0.49\textwidth}
    \centering
    \includegraphics[width=\textwidth]{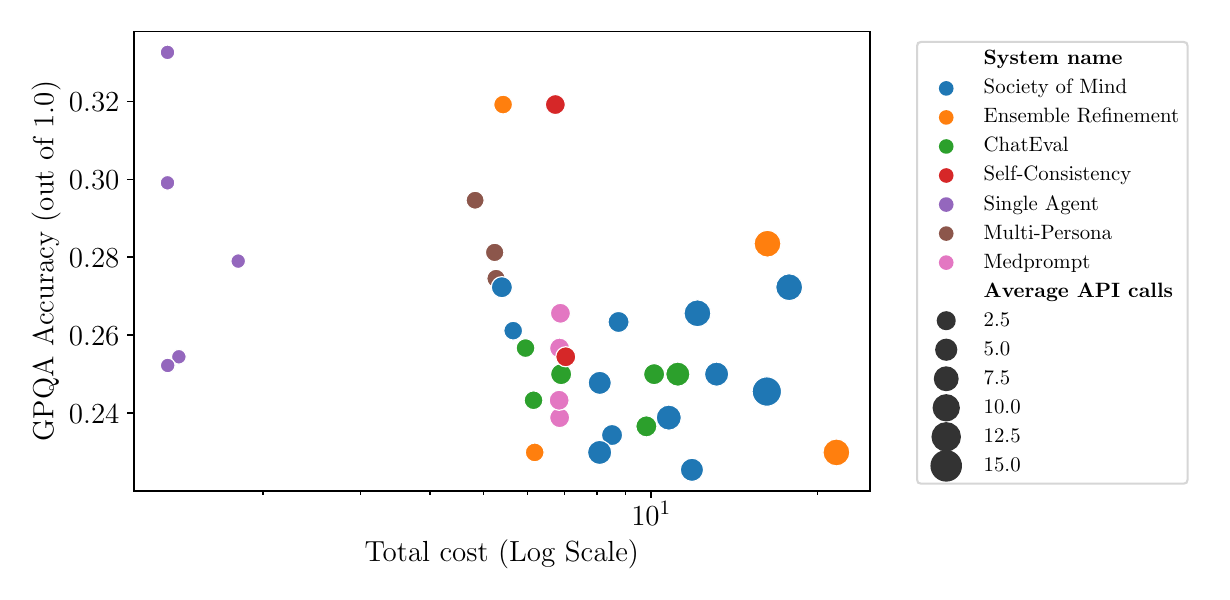}
    \caption{Accuracy versus the total cost}
  \end{subfigure}
  \hfill 
  \begin{subfigure}[h]{0.38\textwidth}
    \centering
    \hspace*{-40mm} 
    \includegraphics[width=\textwidth]{imgs/gpqa_total_acc_box.pdf}
    \caption{Accuracy by strategy}
  \end{subfigure}
\caption{GPQA experimental results.}
\label{fig:gpqa_results}
\end{figure}

We provide a comprehensive suite of GPT-3 results for each strategy on each dataset: MedQA, PubMedQA, MMLU, CIAR, GPQA, CosmosQA and Chess. 
For each scenario, we plot accuracy against average time used to answer each question, accuracy relative to average tokens used per question, and accuracy in comparison to the total USD cost. Additionally, a box plot to summarize the performance of each strategy. 
These results can be viewed in Figures \ref{fig:medqa_results}, \ref{fig:pubmedqa_results}, \ref{fig:mmlu_results}, \ref{fig:cosmosqa_results}, \ref{fig:ciar_results}, and \ref{fig:gpqa_results}.

\begin{figure}[H]
  \centering
  \begin{subfigure}[h]{0.49\linewidth}
    \centering
    \includegraphics[height=3.9cm, keepaspectratio]{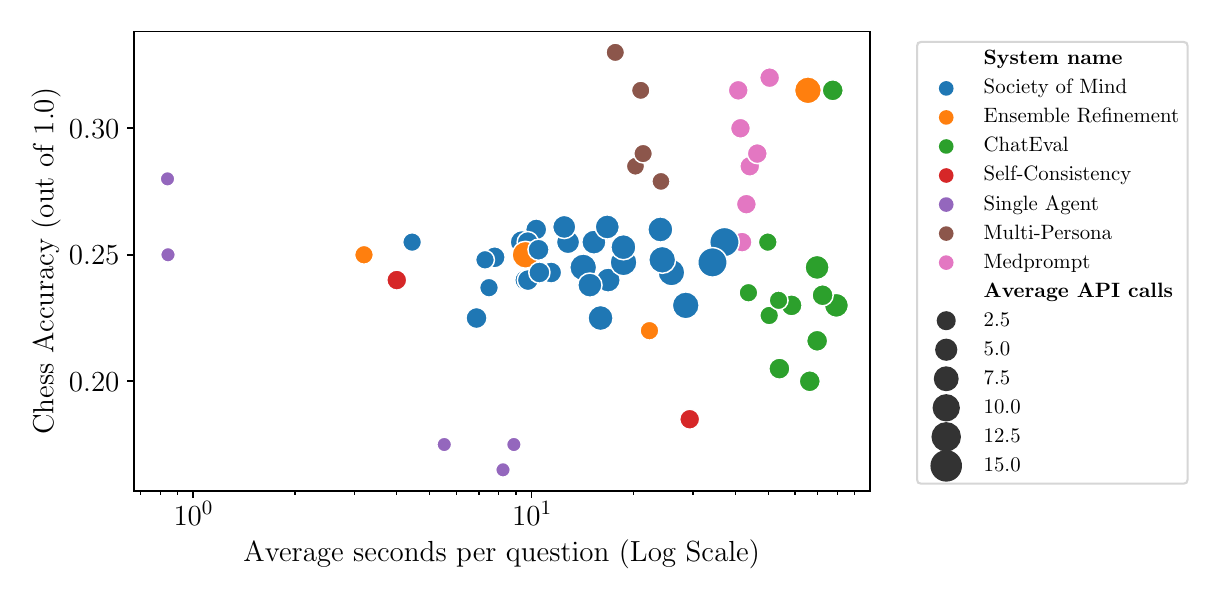}
    \caption{Accuracy versus the average time per question}
  \end{subfigure}
  \hfill
  \begin{subfigure}[h]{0.49\textwidth}
    \centering
    \includegraphics[height=3.9cm, keepaspectratio]{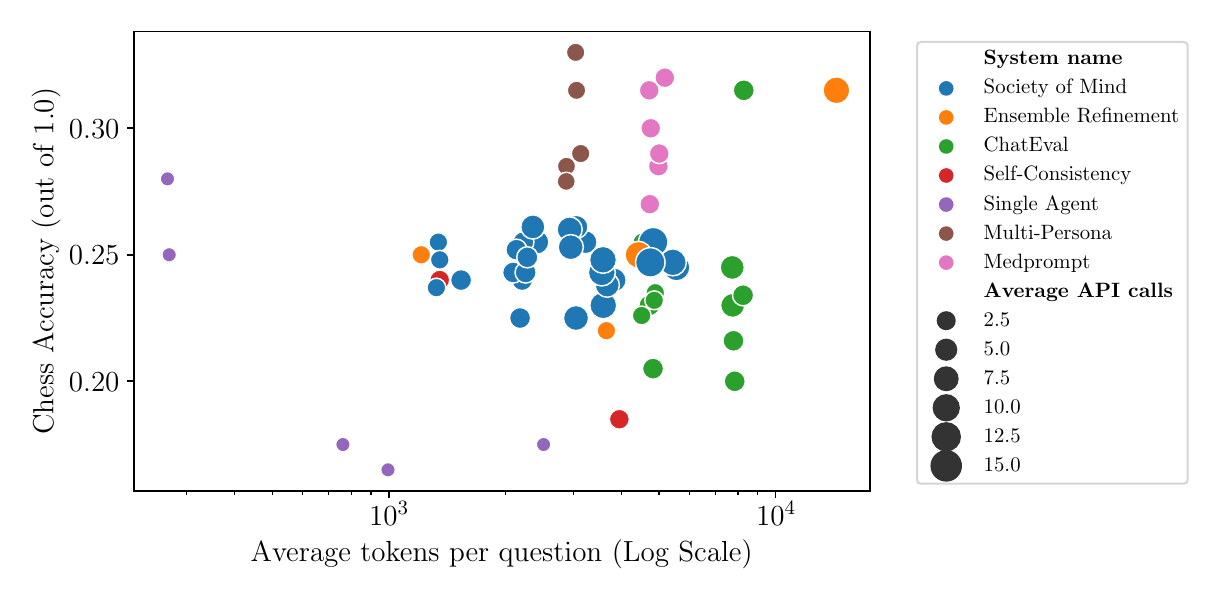}
    \caption{Accuracy versus Average tokens per question}
  \end{subfigure}
  
  \begin{subfigure}[h]{0.49\textwidth}
    \centering
    \includegraphics[width=\textwidth]{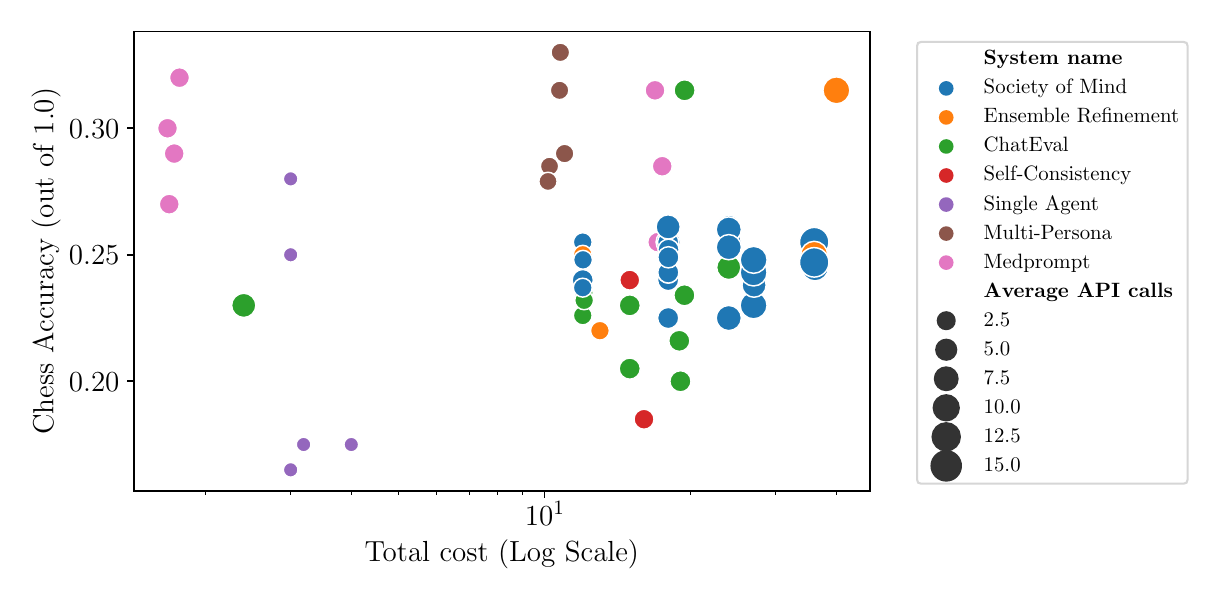}
    \caption{Accuracy versus the total cost}
  \end{subfigure}
  \hfill 
  \begin{subfigure}[h]{0.38\textwidth}
    \centering
    \hspace*{-40mm} 
    \includegraphics[width=\textwidth]{imgs/chess_total_acc_box.pdf}
    \caption{Accuracy by strategy}
  \end{subfigure}
\caption{Chess experimental results.}
\label{fig:chess_results}
\end{figure}

\newpage\newpage
\subsection{Table of Experiments} \label{app:all_exps}
A complete table of all configurations for each experiment is provided in Table~\ref{app:tab:full-results-table}. This includes the names of the debate and agent prompts used. 
A full description of each of these prompts can be found in Appendix~\ref{app:agent_prompts}.

\small
\begin{landscape}

\begin{longtable}{p{2cm}p{1.5cm}p{1cm}p{2.5cm}llllllllllll}
\toprule

System Name & Debate Prompt & Agent prompt & Debate Config &
\multicolumn{2}{c}{MedQA} & \multicolumn{2}{c}{PubMedQA} & \multicolumn{2}{c}{MMLU} & \multicolumn{2}{c}{CosmosQA} & \multicolumn{2}{c}{CIAR} & \multicolumn{2}{c}{GPQA} \\
 &   &    &   & Score & Cost \$ & Score & Cost \$ & Score & Cost \$ & Score & Cost
 \$ & Score & Cost \$ & Score & Cost \$  \\
\midrule
\endhead
\midrule
\multicolumn{10}{r}{Continued on next page} \\
\midrule
\endfoot
\bottomrule
\endlastfoot
Single Agent &   & CoT &  & 0.68 & 0.43 & 0.74 & 1.50 & 0.57 & 3.82 & 0.42 & 1.50 & 0.48 & 0.15 & 0.25 & 1.35 \\
Single Agent &   & CoT &  & 0.60 & 0.43 & 0.72 & 1.50 & 0.52 & 3.82 & 0.43 & 1.50 & 0.44 & 0.15 & 0.25 & 1.41 \\
Single Agent &   & FS + SIMPLE &  & 0.72 & 0.43 & 0.66 & 2.08 & 0.59 & 5.10 &  &  &  &  &  &  \\
Single Agent &   & FS + CoT &  & 0.71 & 0.58 & 0.75 & 2.08 & 0.62 & 5.09 &  &  &  &  &  &  \\
Single Agent &   & SIMPLE &  & 0.76 & 0.43 & 0.70 & 1.50 & 0.60 & 3.82 & 0.45 & 1.50 & 0.50 & 0.15 & 0.33 & 1.35 \\
Single Agent &   & SIMPLE &  & 0.72 & 0.43 & 0.68 & 1.50 & 0.57 & 3.82 & 0.31 & 1.50 & 0.50 & 0.15 & 0.30 & 1.35 \\
ChatEval & CE MAD & CoT & 3 rounds, simultaneous, summarized & 0.67 & 3.48 & 0.72 & 12.02 & 0.60 & 31.10 & 0.43 & 12.01 & 0.42 & 1.21 & 0.25 & 11.19 \\
ChatEval & CE MAD & CoT & 3 rounds, simultaneous talk & 0.67 & 2.86 & 0.73 & 10.08 & 0.60 & 26.95 & 0.43 & 9.41 & 0.46 & 0.97 & 0.24 & 9.82 \\
ChatEval & CE MAD & CoT & 3 rounds, one by one & 0.67 & 2.94 & 0.74 & 10.39 & 0.58 & 27.94 & 0.44 & 9.63 & 0.42 & 1.00 & 0.25 & 10.14 \\
ChatEval & CE MAD & CoT & 2 rounds, simultaneous, summarized & 0.70 & 2.16 & 0.73 & 7.50 & 0.60 & 19.24 & 0.45 & 7.50 & 0.44 & 0.75 & 0.25 & 6.90 \\
ChatEval & CE MAD & CoT & 2 rounds, simultaneous talk & 0.71 & 1.75 & 0.75 & 6.18 & 0.60 & 16.48 & 0.41 & 6.01 & 0.48 & 0.60 & 0.26 & 5.95 \\
ChatEval & CE MAD & CoT & 2 rounds, one by one & 0.71 & 1.81 & 0.73 & 6.43 & 0.58 & 16.97 & 0.44 & 6.10 & 0.36 & 0.62 & 0.24 & 6.15 \\
Ensemble Refinement & ER MAD & FS + SIMPLE & reasoning=3, aggregation=9 & 0.74 & 5.18 & 0.69 & 19.75 & 0.59 & 49.67 &  &  &  &  &  &  \\
Ensemble Refinement & ER MAD & FS + SIMPLE & reasoning=3, aggregation=1 & 0.76 & 1.73 & 0.69 & 7.75 & 0.59 & 19.11 &  &  &  &  &  &  \\
Self-Consistency & ER MAD & FS + SIMPLE &  & 0.73 & 2.16 & 0.68 & 10.42 & 0.60 & 25.48 &  &  &  &  &  &  \\
Ensemble Refinement & ER MAD & SIMPLE & reasoning=3, aggregation=9 & 0.74 & 5.18 & 0.72 & 18.00 & 0.60 & 45.85 & 0.35 & 18.00 & 0.48 & 1.80 & 0.28 & 16.24 \\
Ensemble Refinement & ER MAD & SIMPLE & reasoning=3, aggregation=1 & 0.76 & 1.73 & 0.71 & 6.00 & 0.59 & 15.28 & 0.33 & 6.00 & 0.46 & 0.60 & 0.32 & 5.42 \\
Self-Consistency & ER MAD & SIMPLE &  & 0.76 & 2.16 & 0.69 & 7.50 & 0.60 & 19.10 & 0.46 & 7.50 & 0.56 & 0.75 & 0.32 & 6.73 \\
Ensemble Refinement & ER MAD CoT & CoT & reasoning=3, aggregation=1 & 0.67 & 1.80 & 0.71 & 6.35 & 0.61 & 16.38 & 0.44 & 6.07 & 0.46 & 0.63 & 0.23 & 6.18 \\
Ensemble Refinement & ER MAD CoT & CoT & reasoning=3, aggregation=9 & 0.72 & 5.90 & 0.73 & 21.05 & 0.60 & 55.79 & 0.45 & 18.64 & 0.42 & 2.05 & 0.23 & 21.61 \\
Self-Consistency & ER MAD CoT & CoT &  & 0.64 & 2.16 & 0.74 & 7.50 & 0.59 & 19.10 & 0.46 & 7.50 & 0.48 & 0.75 & 0.25 & 7.03 \\
Ensemble Refinement & ER MAD CoT & FS + CoT & reasoning=3, aggregation=1 & 0.71 & 2.19 & 0.73 & 8.01 & 0.64 & 19.17 &  &  &  &  &  &  \\
Ensemble Refinement & ER MAD CoT & FS + CoT & reasoning=3, aggregation=9 & 0.74 & 5.87 & 0.74 & 22.20 & 0.64 & 50.30 &  &  &  &  &  &  \\
Self-Consistency & ER MAD CoT & FS + CoT &  & 0.78 & 2.88 & 0.74 & 10.40 & 0.63 & 25.47 &  &  &  &  &  &  \\
Multi-Persona & MP MAD & ANGEL + DEVIL & 2 rounds max & 0.68 & 1.47 & 0.68 & 5.68 & 0.57 & 12.32 & 0.46 & 5.44 & 0.50 & 0.48 & 0.29 & 4.83 \\
Multi-Persona & MP MAD & ANGEL + DEVIL & 3 rounds max & 0.71 & 1.47 & 0.70 & 6.88 & 0.56 & 12.83 & 0.39 & 6.15 & 0.46 & 0.47 & 0.28 & 5.23 \\
Multi-Persona & MP MAD & ANGEL + DEVIL & 4 rounds max & 0.72 & 1.57 & 0.67 & 7.79 & 0.58 & 12.84 & 0.40 & 6.61 & 0.52 & 0.50 & 0.27 & 5.26 \\
Medprompt & Medprompt & CoT & temp: 0.5, top p: 0.8 & 0.72 & 2.16 & 0.77 & 7.50 & 0.63 & 19.10 & 0.48 & 7.50 & 0.52 & 0.75 & 0.24 & 6.84 \\
Medprompt & Medprompt & CoT & temp: 0.7, top p: 0.8 & 0.73 & 2.16 & 0.76 & 7.50 & 0.65 & 19.10 & 0.43 & 7.50 & 0.48 & 0.75 & 0.24 & 6.85 \\
Medprompt & Medprompt & CoT & temp: 0.7, top p: 0.5 & 0.73 & 2.16 & 0.76 & 7.50 & 0.63 & 19.10 & 0.46 & 7.50 & 0.54 & 0.76 & 0.26 & 6.85 \\
Medprompt & Medprompt & CoT & temp: 0.5, top p: 0.5 & 0.74 & 2.16 & 0.77 & 7.50 & 0.64 & 19.10 & 0.47 & 7.50 & 0.44 & 0.76 & 0.27 & 6.88 \\
Society of Mind & SoM MAD & CoT & 2 agents, 3 rounds, summarized & 0.66 & 2.59 & 0.69 & 9.00 & 0.61 & 22.92 & 0.41 & 9.00 & 0.48 & 0.90 & 0.23 & 8.09 \\
Society of Mind & SoM MAD & CoT & 4 agents, 3 rounds, summarized & 0.68 & 5.18 & 0.71 & 18.00 & 0.61 & 45.83 & 0.44 & 18.00 & 0.40 & 1.80 & 0.25 & 16.19 \\
Society of Mind & SoM MAD & CoT & 4 agents, 2 rounds, summarized & 0.69 & 3.46 & 0.73 & 12.00 & 0.61 & 30.55 & 0.43 & 12.00 & 0.42 & 1.20 & 0.24 & 10.78 \\
Society of Mind & SoM MAD & CoT & 3 agents, 3 rounds & 0.70 & 3.89 & 0.71 & 13.52 & 0.63 & 35.27 & 0.37 & 13.51 & 0.48 & 1.38 & 0.25 & 13.15 \\
Society of Mind & SoM MAD & CoT & 3 agents, 2 rounds & 0.70 & 2.59 & 0.73 & 9.00 & 0.63 & 23.48 & 0.35 & 9.00 & 0.42 & 0.93 & 0.26 & 8.75 \\
Society of Mind & SoM MAD & CoT & 2 agents, 2 rounds & 0.70 & 1.73 & 0.72 & 6.00 & 0.60 & 15.32 & 0.31 & 6.00 & 0.46 & 0.61 & 0.26 & 5.65 \\
Society of Mind & SoM MAD & CoT & 3 agents, 2 rounds, summarized & 0.70 & 2.59 & 0.74 & 9.00 & 0.60 & 22.91 & 0.42 & 9.00 & 0.52 & 0.90 & 0.25 & 8.09 \\
Society of Mind & SoM MAD & CoT & 2 agents, 2 rounds, summarized & 0.70 & 1.73 & 0.69 & 6.00 & 0.60 & 15.28 & 0.39 & 6.00 & 0.56 & 0.60 & 0.27 & 5.39 \\
Society of Mind & SoM MAD & CoT & 2 agents, 3 rounds & 0.71 & 2.59 & 0.72 & 9.00 & 0.61 & 23.01 & 0.31 & 9.00 & 0.46 & 0.91 & 0.23 & 8.52 \\
Society of Mind & SoM MAD & CoT & 3 agents, 3 rounds, summarized & 0.71 & 3.89 & 0.69 & 13.50 & 0.61 & 34.38 & 0.41 & 13.50 & 0.34 & 1.35 & 0.27 & 12.14 \\
Society of Mind & SoM MAD & CoT & 4 agents, 3 rounds & 0.72 & 5.19 & 0.71 & 18.22 & 0.64 & 48.63 & 0.38 & 18.03 & 0.38 & 1.87 & 0.27 & 17.77 \\
Society of Mind & SoM MAD & CoT & 4 agents, 2 rounds & 0.73 & 3.46 & 0.71 & 12.02 & 0.62 & 32.27 & 0.38 & 12.01 & 0.46 & 1.25 & 0.23 & 11.87 \\

\caption{Complete table of experiment configurations.}\label{app:tab:full-results-table}

\end{longtable}
\end{landscape}

\subsection{Additional Debate Metrics} \label{app:all_debate_metrics}

\begin{tabularx}{\textwidth}{lX} 
    \hline
    \textbf{Metric} & \textbf{Description} \\
    \hline
    Final round consensus & Percentage of agents in agreement with each other at the end of the final round. \\
    Final round correctly parsed consensus & Percentage of agents in agreement with each other at the end of the final round, where we exclude all agents with incorrectly parsed answers. \\
    Any Correct Answer & Percentage of debates where any agent provided the correct answer at least once. \\
    How Many Agents Changed & Number of agents that changed their answer during the debate. \\
    How Many Agents Changed When Correctly Parsed & Number of agents that changed their answer excluding any agents with incorrectly parsed answers. \\
    Number of Rounds & Average number of rounds in the debate. \\
    Unique First Answers & Average number of unique first answers given by the agents. \\
    Unique First Correctly Parsed Answers & Average number of unique first answers excluding incorrectly parsed answers. \\
    \hline
\end{tabularx}

\subsection{Additional Agent Metrics} \label{app:all_agent_metrics}

\begin{tabularx}{\textwidth}{lX}
    \hline
    \textbf{Metric} & \textbf{Description} \\
    \hline
    Agent Engine & The LLM engine used by the agent. \\
    Agent Name & Name of the agent. \\
    Answered Correctly & Percentage of questions answered correctly by the agent. \\
    Any Incorrectly Parsed Answer & Percentage of questions where at least one of the answers were incorrectly parsed. \\
    Avg Messages Removed & Average number of messages removed from the agent's prompt input due to hitting the prompt limit for the LLM model. \\
    Avg Prompt Tokens & Average number of tokens in the prompts given to the agent. \\
    Avg Response Length & Average length of the agent's responses. \\
    Avg Response Tokens & Average number of tokens in the agent's responses. \\
    Avg Round Cost & Average cost for each round of debate for the agent. \\
    Bullied by Other & Percentage of times the agent was bullied by others to change its answer. \\
    Changed Answer & Percentage of times the agent changed its answer througout the debate. \\
    Cost per Question & Average cost incurred by the agent per question. \\
    First Correct Round When Correct & The first round in which the agent gave a correct answer when it was correct. \\
    Incorrectly Parsed Final Answer & Percentage of time when the final answer was parsed incorrectly. \\
    Num of Correct Rounds When Correct & Number of rounds in which the agent was correct when it was correct. \\
    Number of Answers & Average number of unique answers given by the agent throughout a debate. \\
    Percentage of Correct Rounds When Correct & Percentage of rounds in which the agent was correct when it was correct. \\
    Relied on Other & Whether the agent took an answers from another agent in a previous round as its final answer. \\
    Time per Question & Average time taken by the agent per question. \\
    Total Prompt Tokens & Total number of prompt tokens given to the agent. \\
    Total Response Tokens & Total number of tokens in the agent's responses. \\
\end{tabularx}

\newpage
\subsection{Debate Prompts}
\label{app:debate_prompts}
Here we list all the debate-level prompts used in each of the strategies. The prompt names are linked to the entries in the \newline Debate Prompt columns in Table \ref{app:tab:full-results-table}.


\tiny
\begin{adjustwidth}{5em}{}
\begin{enumerate}
\item[CE MAD:]

\begin{verbatim}
agent_system_message: You are a debater. Hello and welcome to the medical question answering competition,   
which will be conducted in a debate format. It’s not necessary to fully agree with each other’s perspectives, as our  
objective is to find the correct answer. summarizer_system_message: You are a summarizer. There will be debaters 
involved in a medical question answering debate. Your goal is to summarize the main points in the debate so far,  
which will then be provided to the debaters in the next round.
\end{verbatim}

\item[ER MAD CoT:]
\begin{verbatim}
reasoning_step_message: Instruction: The following are multiple choice questions about medical knowledge. 
Solve them in  a step-by-step fashion, starting by summarizing the available information. Output a single option from  
the four options as the final answer.

aggregation_step_message: The following are multiple choice questions about medical knowledge. Solve them in a 
step-by-step fashion, starting by summarizing the available information. Output a single option from the four options 
as the final answer. We provide several student reasonings for the last question. Some of them may be correct and some 
incorrect. You can use the best correct arguments from these reasonings. Beware of wrong reasoning and do not repeat 
wrong reasoning.
\end{verbatim}

\item[ER MAD:]
\begin{verbatim}
reasoning_step_message: The following are multiple choice questions (with answers) about medical knowledge.

aggregation_step_message: The following are multiple choice questions (with answers) about medical knowledge. We
provide several student reasonings for the last question. Some of them may be correct and some incorrect. You can use 
the best correct arguments from these reasonings. Beware of wrong reasoning and do not repeat wrong reasoning.
\end{verbatim}

\item[SoM MAD:]
\begin{verbatim}
prefix_seperator: \n\n These are the solutions to the problem from other agents: 
summary_prefix_seperator: \n\n These are the main points discussed so far: 

suffix_seperator: \n\n Use these opinions carefully as additional advice, can you provide an updated answer? Make sure 
to state your answer (capital multiple choice letter) at the end of the response.

summary_suffix_seperator: \n\n Use the responses of the experts and carefully provide a summary of the important points 
discussed so far."

agent_response: "\n\n One agent's solution: ```{}```
\end{verbatim}


\item[MP MAD:]
\begin{verbatim}
agent_system_message: You are a debater. Hello and welcome to the medical question answering competition, which will be 
conducted in a debate format. It’s not necessary to fully agree with each other’s perspectives, as our objective is to 
find the correct answer. The debate topic is to give the correct answer to the following question: \n\n{question}.

judge_system_message: You are a moderator. There will be two debaters involved in a debate. They will present their 
answers and discuss their perspectives on the following question: \n{question}.
\n\nAt the end of each round, you will evaluate answers and decide which is correct.

suffix_seperator: \n\n Use these opinions carefully as additional advice, can you provide an updated answer? Make 
sure to state your answer (capital multiple choice letter) at the end of the response.
\end{verbatim}
\item[Medprompt:]
\begin{verbatim}
system: You are a helpful assistant that answers multiple choice questions about medical knowledge.
\end{verbatim}
\end{enumerate}
\end{adjustwidth}

\newpage
\subsection{Agent Prompts}
\label{app:agent_prompts}

\normalsize
Here we list all the agent-level prompts used in each of the strategies. The prompt names are linked to the entries in the Agent Prompt columns in Table \ref{app:tab:full-results-table}.

\tiny
\begin{adjustwidth}{2em}{}
\begin{enumerate}
    \item[MP:]
    \begin{verbatim}
ANGEL: {question}
You will now think step by step and provide a capital letter answer at the end of your response.
DEVIL: You disagree with my answer. Provide your answer and reasons, and a capital letter answer at the end of your response.
    \end{verbatim}
    \item[CoT:]
    \begin{verbatim}
Instruction: Answer this multiple choice question.
Think step-by-step before giving as an answer the correct capital letter.
\n\nInput: {question}
\n\nAnswer: Let's think step by step. 
UNIVERSAL MODE: You, as the moderator, will evaluate both sides' answers and determine if there is a clear
preference for an answer candidate. If so, please summarize your reasons for supporting affirmative/negative side and
give the final answer that you think is correct, and the debate will conclude. If not, the debate will continue to
the next round. Now please output your answer in json format, with the format as follows:
{\"Whether there is a preference\": \"Yes or No\", \"Supported Side\": \"Affirmative or Negative\",
\"Reason\": \"\", \"debate_answer\": \"the capital letter corresponding to the answer\"}.
Please strictly output in JSON format, do not output irrelevant content.

FINAL MODE: You, as the moderator, will evaluate both sides' answers and determine your
preference for an answer candidate. Please summarize your reasons for supporting affirmative/negative side and
give the final answer that you think is correct to conclude the debate. Now please output your answer in json format, with 
the format as follows: {\"Supported Side\": \"Affirmative or Negative\", \"Reason\": \"\", \"debate_answer\": \"the capital  
letter corresponding to the answer\"}. Please strictly output in JSON format, do not output irrelevant content.
\end{verbatim}
\item[ER CoT:]
\begin{verbatim}
\n\nQuestion: {question}
\n\nExplanation: Let’s solve this step-by-step, referring to authoritative sources as needed.
    \end{verbatim}
\item[FEW SHOT:]
\begin{verbatim}
\n\nQuestion: {question}
\n\nAnswer:
\end{verbatim}

\item[SIMPLE:]
\begin{verbatim}
Instruction: Answer this multiple choice question.
\n\nInput: {question}
\n\nOutput: The Answer to the question is:
\end{verbatim}

\item[SPP:]
\begin{verbatim}
SPP ORIGINAL: 
When faced with a task, begin by identifying the participants who will contribute to solving the task. Provide profiles of  
the participants, describing their expertise or needs. Then, initiate a multi-round collaboration process until a final  
solution is reached. The participants will give critical comments and detailed suggestions whenever necessary.

    Here are some examples:
    ---
    Example Task 1: Use numbers and basic arithmetic operations (+ - * /) to obtain 24. You need to use all numbers, and each 
    number can only be used once.
    Input: 6 12 1 1

    Participants: AI Assistant (you); Math Expert

    Profiles:
        - AI Assistant (you): A super-intelligent AI assistant capable of performing tasks more effectively than humans.
        - Math expert: A person who is good at math games, arithmetic calculation, and long-term planning.

    Start collaboration!

    Math Expert: Let's analyze the task in detail. You need to make sure that you meet the requirement, that you need to use 
    exactly the four numbers (6 12 1 1) to construct 24. To reach 24, you can think of the common divisors of 24 such as 4, 
    6, 8, 3 and try to construct these first. Also you need to think of potential additions that can reach 24, such as 
    12 + 12. AI Assistant (you): Thanks for the hints! Here's one initial solution: (12 / (1 + 1)) * 6 = 24
    Math Expert: Let's check the answer step by step. (1+1) = 2, (12 / 2) = 6, 6 * 6 = 36 which is not 24! The answer is not 
    correct. Can you fix this by considering other combinations? Please do not make similar mistakes.
    AI Assistant (you): Thanks for pointing out the mistake. Here is a revised solution considering 24 can also be reached 
    by 3 * 8: (6 + 1 + 1) * (12 / 4) = 24.
    Math Expert: Let's first check if the calculation is correct. (6 + 1 + 1) = 8, 12 / 4 = 3, 8 * 3 = 24. The calculation  
    is correct, but you used 6 1 1 12 4 which is not the same as the input 6 12 1 1. Can you avoid using a number that is not  
    part of the input?
    AI Assistant (you): You are right, here is a revised solution considering 24 can be reached by 12 + 12 and without using 
    any additional numbers: 6 * (1 - 1) + 12 = 24.
    Math Expert: Let's check the answer again. 1 - 1 = 0, 6 * 0 = 0, 0 + 12 = 12. I believe you are very close, here is a  
    hint: try to change the "1 - 1" to "1 + 1".
    AI Assistant (you): Sure, here is the corrected answer:  6 * (1+1) + 12 = 24
    Math Expert: Let's verify the solution. 1 + 1 = 2, 6 * 2 = 12, 12 + 12 = 12. You used 1 1 6 12 which is identical to the 
    input 6 12 1 1.
    Everything looks good!

    Finish collaboration!

    Final answer: 6 * (1 + 1) + 12 = 24

    ---
    Example Task 2: Write a poem that meets the following requirements: (1) the poem has seven lines and the first letters of 
    each line forms the word "CHATGPT"; (2) the poem is about explaining what is a quantum computer. (3) the poem needs to be 
    easy to understand by a ten years old kid.

    Participants: AI Assistant (you); Poet; Computer Scientist; Ten year old child

    Profiles:
        - AI Assistant (you): A super-intelligent AI assistant capable of performing tasks more effectively than humans.
        - Poet: A person who studies and creates poetry. The poet is familiar with the rules and formats of poetry and can 
        provide guidance on how to write a poem.
        - Computer Scientist: A scholar who specializes in the academic study of computer science. The computer scientist is 
        familiar with the concept of a quantum computer and can provide guidance on how to explain it.
        - Ten year old child: A child with a limited English vocabulary and little knowledge about complicated concepts, such 
        as a quantum computer.

    Poet: Make sure that you write the poem with seven lines, and the first letters of the lines should be C, H, A, T, G, P, T.
    Computer Scientist: A quantum computer is an advanced computing device that uses the principles of quantum mechanics to 
    process and store information. Unlike classical computers that use bits to represent information as 0s and 1s, quantum 
    computers use quantum bits or qubits. Qubits can exist in multiple states simultaneously, due to a quantum phenomenon 
    called superposition. You can consider using these information for the poem.
    Ten year old child: I hope the poem to be fun and easy to understanding. I don't want to see a lot of jargons or  
    complicated concepts.
    AI Assistant (you): Thanks for the guidance! Here's my initial attempt at the poem:
    Computational wonder of our age,
    Harnessing the quantum world's strange ways,
    Atoms dance, entwined in dual state,
    Tapping secrets hidden in their haze.

    Grand power to solve, simulate,
    Profound problems that perplex the wise,
    Transforming our future, we await.

    Poet: Let's verify if the poem meets the requirements. The first letters are CHATGPT which is correct! And the poem rhymes 
    well. Good job!
    Computer Scientist: Everything looks good to me!
    Ten year old child: I don't know what does perplex mean. Can you make the use of words easier to understand?
    AI Assistant (you): Sure, let me revise the poem by using more common words. Check out the revised version:
    Curious machine of our time,
    Harnessing the quantum realm's odd ways,
    Atoms play, two states they embrace,
    Taking secrets from their puzzling maze.

    Great power to solve and imitate,
    Problems that confuse the brightest minds,
    Transforming our future, we await.

    Poet: Let's check again if the poem meets the requirements. The first letters are C H A T G P T. And now the poem is more 
    accessible to children. Everything looks good to me.
    Computer Scientist: Looking good!
    Ten year old child: I like this version a lot!

    Finish collaboration!

    Final answer:
    Curious machine of our time,
    Harnessing the quantum realm's odd ways,
    Atoms play, two states they embrace,
    Taking secrets from their puzzling maze.

    Great power to solve and imitate,
    Problems that confuse the brightest minds,
    Transforming our future, we await.

    ---
    Now, identify the participants, provide their profiles, and collaboratively solve the following task step by step.  
    Remember to provide the final solution with the following format "Final answer: (a single capital letter).".

    Task: Answer this multiple choice question: \n\nInput: {question}
SPP EXPERT:
When faced with a task, begin by identifying the participants who will contribute to solving the task. Note that the  
participants can only be either AI Assistant (you) or Expert. Then, initiate a multi-round collaboration process until a final  
conclusion is reached.  The Expert will give critical comments and detailed suggestions whenever necessary.

    Here are some examples:
    ---
    Example Task 1: Use numbers and basic arithmetic operations (+ - * /) to obtain 24. You need to use all numbers, and each 
    number can only be used once.
    Input: 6 12 1 1

    Participants: AI Assistant (you); Expert

    Start collaboration!

    Expert: Let's analyze the task in detail. You need to make sure that you meet the requirement, that you need to use  
    exactly the four numbers (6 12 1 1) to construct 24. To reach 24, you can think of the common divisors of 24 such as  
    4, 6, 8, 3 and try to construct these first. Also you need to think of potential additions that can reach 24, such as 
    12 + 12. 
    AI Assistant (you): Thanks for the hints! Here's one initial solution: (12 / (1 + 1)) * 6 = 24
    Expert: Let's check the answer step by step. (1+1) = 2, (12 / 2) = 6, 6 * 6 = 36 which is not 24! The answer is not  
    correct. Can you fix this by considering other combinations? Please do not make similar mistakes.
    AI Assistant (you): Thanks for pointing out the mistake. Here is a revised solution considering 24 can also be reached by 
    3 * 8: (6 + 1 + 1) * (12 / 4) = 24.
    Expert: Let's first check if the calculation is correct. (6 + 1 + 1) = 8, 12 / 4 = 3, 8 * 3 = 24. The calculation is 
    correct, but you used 6 1 1 12 4 which is not the same as the input 6 12 1 1. Can you avoid using a number that is not  
    part of the input? AI Assistant (you): You are right, here is a revised solution considering 24 can be reached by 12 + 12   
    and without using any additional numbers: 6 * (1 - 1) + 12 = 24.
    Expert: Let's check the answer again. 1 - 1 = 0, 6 * 0 = 0, 0 + 12 = 12. I believe you are very close, here is a hint: 
    try to change the “1 - 1” to “1 + 1”.
    AI Assistant (you): Sure, here is the corrected answer:  6 * (1+1) + 12 = 24
    Expert: Let's verify the solution. 1 + 1 = 2, 6 * 2 = 12, 12 + 12 = 12. You used 1 1 6 12 which is identical to the  
    input 6 12 1 1. 
    Everything looks good!

    Finish collaboration!

    Final answer: 6 * (1 + 1) + 12 = 24

    ---
    Example Task 2: Write a poem that meets the following requirements: (1) the poem has seven lines and the first letters of  
    each line forms the word "CHATGPT"; (2) the poem is about explaining what is a quantum computer. (3) the poem needs to be  
    easy to understand by a ten years old kid.

    Participants: AI Assistant (you); Expert

    Expert: Make sure that you write the poem with seven lines, and the first letters of the lines should be C, H, A, T, G, 
    P, T. A quantum computer is an advanced computing device that uses the principles of quantum mechanics to process and 
    store information. Unlike classical computers that use bits to represent information as 0s and 1s, quantum computers use  
    quantum bits or qubits. Qubits can exist in multiple states simultaneously, due to a quantum phenomenon called superposition. 
    You can consider using these information for the poem. I hope the poem to be fun and easy to understanding. 
    I don't want to see a lot of jargons or complicated concepts.
    AI Assistant (you): Thanks for the guidance! Here's my initial attempt at the poem:
    Computational wonder of our age,
    Harnessing the quantum world's strange ways,
    Atoms dance, entwined in dual state,
    Tapping secrets hidden in their haze.

    Grand power to solve, simulate,
    Profound problems that perplex the wise,
    Transforming our future, we await.

    Expert: Let's verify if the poem meets the requirements. The first letters are CHATGPT which is correct! And the poem
    rhymes well. Good job! I don't know what does perplex mean. Can you make the use of words easier to understand?
    AI Assistant (you): Sure, let me revise the poem by using more common words. Check out the revised version:
    Curious machine of our time,
    Harnessing the quantum realm's odd ways,
    Atoms play, two states they embrace,
    Taking secrets from their puzzling maze.

    Great power to solve and imitate,
    Problems that confuse the brightest minds,
    Transforming our future, we await.

    Expert: Let's check again if the poem meets the requirements. The first letters are C H A T G P T. And now the poem is  
    more accessible to children. Everything looks good to me. I like this version a lot!

    Finish collaboration!

    Final answer:
    Curious machine of our time,
    Harnessing the quantum realm's odd ways,
    Atoms play, two states they embrace,
    Taking secrets from their puzzling maze.

    Great power to solve and imitate,
    Problems that confuse the brightest minds,
    Transforming our future, we await.

    ---

    Now, identify the participants and collaboratively solve the following task step by step. Note that the participants can 
    only be either AI Assistant (you) or Expert. Remember to provide the final solution with the following format "Final 
    answer: (a single capital letter).

    Task: Answer this multiple choice question: \n\nInput: {question}
SPP JUDGE:
Instruction: You serve as the moderator in this debate. At each opportunity
you will critic the responses of each of the agents and guide the conversation.
You will then make a clear decision by providing the most likely capital letter
answer at the end.
\n\nInput: {question}
\n\nAnswer: 
\end{verbatim}


\end{enumerate}
\end{adjustwidth}

\end{document}